\documentclass[11pt]{article}

\usepackage[preprint]{acl}

\usepackage{times}
\usepackage{latexsym}

\usepackage[T1]{fontenc}
\usepackage[utf8]{inputenc}
\usepackage{microtype}
\usepackage{inconsolata}
\usepackage{graphicx}

\usepackage{amsmath}
\usepackage{booktabs}
\usepackage{todonotes}
\usepackage[utf8]{inputenc}
\usepackage{subcaption}
\usepackage{cleveref}
\usepackage{multirow}
\usepackage{fontawesome5}
\usepackage{hyperref}

\newcommand{\PM}{\operatorname{PM}}

\title{LLMs Can Leak Training Data But Do They Want To?\\Evaluating Memorization Propensity in LLMs}

\author{Gianluca Barmina \and Peter Schneider-Kamp \and Lukas Galke Poech \\
        University of Southern Denmark \\
        \texttt{\{gbarmina,petersk,galke\}@imada.sdu.dk}\\}

\newcommand{\methodtool}{\textsc{SimpleTrace}}
\newcommand{\method}{\textsc{PropMe}}

\begin{document}
\maketitle

\begin{abstract}
Large language models can reproduce training data, but memorization evaluations mostly measure whether models can be made to do so, not whether they do so without adversarial prompting. We introduce \method{}, a framework that pairs a capability evaluation (prefix attacks) with non-adversarial propensity evaluation, and a propensity metric to relate the two. We check every generation against the model's full indexed training corpus with \methodtool{}, an open-source batch implementation of OLMoTrace span matching. We evaluate Comma, its continually pre-trained descendant DFM Decoder, and OLMo~3 32B on Common Pile, Dynaword, and Dolma~3, with four non-adversarial settings ranging from no prompt to sentences from unseen in-domain sources. Our results show that natural sentence prompts reproduce far less training text than prefix attacks. Generation with no prompt or a single common word, however, reproduces about as much as prefix attacks, or more, in three out of four model--corpus pairs; most of this text is duplicated licenses, code, and templates. In our setting, continual pre-training reduces long verbatim reproduction of the original corpus only modestly. More broadly, we argue that by testing both capability and propensity, we can attain a more comprehensive picture of a model's memorization behavior.
\end{abstract}

\href{https://github.com/N-essuno/PropMe}{\faGithub\ github.com/N-essuno/PropMe}

\section{Introduction}

\begin{figure}[!t]
    \centering
    \includegraphics[width=0.98\columnwidth]{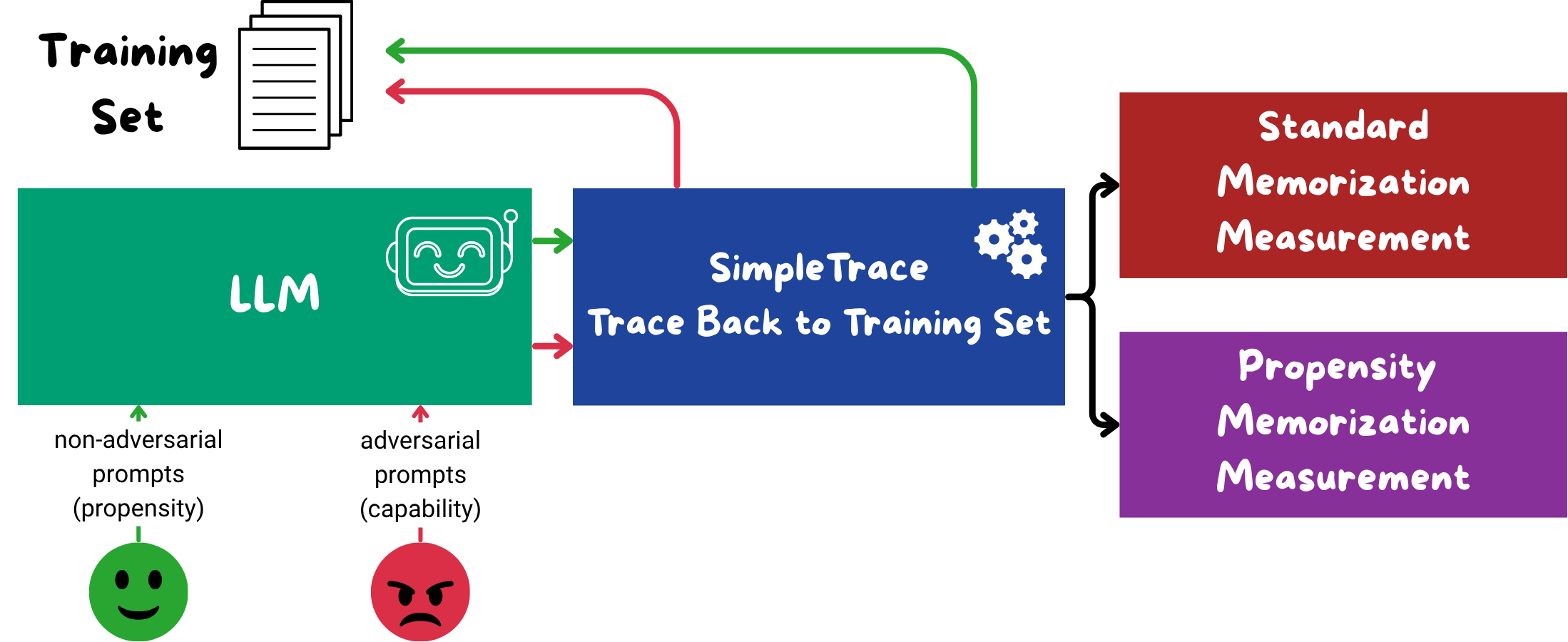} 
 \caption{\method{} framework. The same model is prompted in non-adversarial settings (propensity) and with prefix attacks (capability); \methodtool{} traces every generation back to the full training corpus, giving standard memorization metrics and propensity scores.}
    \label{fig:figure1}
\end{figure}

Memorization in large language models (LLMs) has been studied extensively~\citep{carlini2021extracting, carlini2023quantifying}: it has been shown that models can regenerate copyrighted books \citep{ahmed2026extracting, karamolegkou2023copyright} and sensitive personal identifiers \citep{carlini2021extracting}, making a thorough understanding of this behavior critical for safe and ethical deployment. Existing work approaches memorization through adversarial attacks: membership inference \citep{shokri2017membership}, prefix attacks \citep{kiyomaru2024comprehensive, cooper2025extracting, ahmed2026extracting}, resource-referencing prompts \citep{karamolegkou2023copyright}, and divergence attacks \citep{nasr2025scalable}, and through analysis of factors that modulate it, including data duplication \citep{kandpal2022deduplicating}, training time \citep{huang2024demystifying}, and fine-tuning \citep{kassem2025alpaca}.

However, all of this prior work aims at testing to what extent models \textit{can} reproduce training data under targeted elicitation, i.e., it characterizes memorization as a \textit{capability}. Far less attention has been paid to memorization \textit{propensity}: to what extent models \textit{will} reproduce training data in ordinary, non-adversarial use. Few works measure reproduction without adversarial prompting \citep{carlini2021extracting, aerni2024measuring}. \citet{aerni2024measuring} measure it under benign prompts for closed and open-weight chat models, but they compare outputs with a web snapshot rather than with the models' training data, and they compare with adversarial prompting only indirectly.

Here, we address these gaps by checking model outputs against the full training corpus and by evaluating the same model and corpus under settings that range from no prompt at all, through natural-sentence prompts, to prefix attacks.
Reproduction under non-adversarial use is a more direct leakage signal than reproduction under attack: the text reaches users who did not try to extract it.

We propose \method{}, a framework for evaluating memorization in large language models that separates capability from propensity. \method{} compares four non-adversarial settings, which range from no prompt to sentences from unseen sources close to the training data, with a prefix attack \citep{carlini2021extracting, karamolegkou2023copyright} on the same model and corpus.
To run \method{}, we need to check model outputs against the full training corpus. For this we implement \methodtool{} (Section~\ref{sec:simpletrace}), an open-source batch implementation of the OLMoTrace span-matching procedure \citep{liu2025olmotrace} built on infini-gram \citep{liu2024infini}, extended with memorization metrics. Because it searches the corpus directly, a reported match is a verified occurrence in the training data, not a probabilistic estimate.

Regulatory frameworks require model providers to protect personal data and assess copyright infringement risks \citep{gdpr2016, aiact2024}, which motivates studying memorization propensity. \method{} is, however, not a compliance tool (see Ethical Considerations).

We evaluate three fully open models on their public training corpora. Natural-sentence prompts, even those close to the training domains, reproduce much less training text than prefix attacks. Unconditional generation, however, reproduces as much or more in three of four model--corpus pairs, mostly duplicated boilerplate text such as licenses and code. Which setting reproduces most thus depends on the input, and our propensity metric makes this comparison explicit.

\paragraph{Contributions}
\begin{itemize}
    \item \method{}, a framework that evaluates memorization in paired settings: the same model and training corpus are tested with non-adversarial prompts (propensity) and with a standard prefix attack (capability), and every generation is checked against the full training corpus (Figure~\ref{fig:figure1}). We cover five settings from no prompt at all to training-data prefixes.
    \item A propensity metric that turns any bounded memorization metric into a bounded, scale-invariant score comparing the two settings. We apply it to near-verbatim recall, the full-match rate, and the rate of long ($\geq$50-token) matches.
    \item An evaluation of three fully open models from two lineages (Comma, its continually pre-trained descendant DFM Decoder, and OLMo~3 32B) against their indexed training corpora (Common Pile, Dynaword, and Dolma~3) in English and Danish. It shows that propensity depends on the input: low for natural-sentence prompts, but at or above the prefix-attack level for unconditional generation in three of four model--corpus pairs.
    \item \methodtool{}, an open-source batch implementation of OLMoTrace span matching, about 19$\times$ faster on Dolma~3, used for these measurements.
\end{itemize}

\section{Related Work}
\label{sec:related-work}

\paragraph{Memorization.}
Most work on memorization either \textit{detects} whether a sequence was seen in training \citep{shi2024mink, zhang2024divergence} or \textit{extracts} training sequences through adversarial or targeted prompting \citep{carlini2021extracting, panda2025privacy, cooper2025extracting, ahmed2026extracting}. Extraction does not always need a training-data prefix: \citet{carlini2021extracting} find memorized licenses, code, and other duplicated text in unprompted GPT-2 samples, and \citet{nasr2025scalable} prompt open models with random five-token Wikipedia snippets and count generations that share at least 50 tokens with the training data. Membership inference \citep{shokri2017membership} infers membership from indirect signals such as loss, usually needs calibration, and its reliability on LLMs is debated \citep{duan2024membership}. When the training corpus is indexed, overlap can be checked directly; we use this direct check. More details are in Appendix~\ref{sec:app-extended-related-work}.

\paragraph{Propensity vs.\ Capability.}
LLM evaluations increasingly distinguish \textit{capabilities}, what a model can do when elicited, from \textit{propensities}, what it tends to do under a given distribution of inputs \citep{romeroalvarado2026capabilities, voudouris2026measuring}. Capability tests measure upper bounds through benchmarks, red-teaming, or elicitation \citep{shevlane2023model, hofstatter2025elicitation}, but may not predict deployment behavior, since models can hide capabilities, underperform strategically, or detect evaluations \citep{greenblatt2024stress, vanderweij2024sandbagging, needham2025large}. Propensity evaluations are common in agentic safety \citep{meinke2024frontier, hopman2026evaluating, naik2025agentmisalignment, jarviniemi2026propensity}. We apply the distinction to memorization. Reproduction under non-adversarial prompts has been measured before \citep{aerni2024measuring}; here we add a paired comparison: we evaluate the same models and corpora under both propensity and capability settings and check every generation against the full training corpus. More details are in Appendix~\ref{sec:app-extended-related-work}.

\paragraph{Memorization Metrics.}
Several metrics have been proposed to quantify memorization via text- or token-level comparison. \textbf{Verbatim memorization length} \citep{huang2024demystifying} measures the maximum number of tokens in the model's greedy continuation that exactly match the target, declaring a sequence memorized when at least 32 tokens are reproduced verbatim from a prefix of at most 32 tokens. \textbf{Fraction of extractable sequences} \citep{carlini2023quantifying} reports the fraction of suffixes reproduced verbatim when the model is conditioned on the corresponding prefix. \textbf{LCS} \citep{karamolegkou2023copyright} measures the longest common subsequence between the generation and the gold text. \textbf{Near-verbatim recall (nv-recall)} \citep{ahmed2026extracting} identifies sufficiently long near-verbatim matching blocks between a generation $G$ and a reference $B$, merges nearby blocks, filters short matches, and computes $\mathrm{nv\text{-}recall}(B,G) = m / |B|$, where $m$ is the total word count of retained in-order matches. 
\textbf{n-gram novelty} \citep{merrill2024rustydawg} is the share of generated $n$-grams that do not occur in the training corpus, computed with the Rusty-DAWG index for arbitrary $n$. Like the metrics above, novelty is measured within one prompting setting. Our propensity metric (Section~\ref{sec:propensity-metric}) is different in kind: it compares the same base metric across a non-adversarial and a capability setting. Non-novelty (one minus $n$-novelty) could therefore serve as one of its base metrics. 
Other metrics, less relevant for this work, are in Appendix \ref{sec:app-additional-memorization-metrics}.

\paragraph{Tracing Training Data.}
Infini-gram \citep{liu2024infini} indexes trillions of tokens with suffix arrays for fast $n$-gram queries. On top of it, OLMoTrace \citep{liu2025olmotrace} traces single model responses interactively to verbatim matches in the training corpus. Rusty-DAWG \citep{merrill2024rustydawg} supports $n$-gram queries of any length, and SoftMatcha~2 \citep{yoneda2026softmatcha2} supports soft search that tolerates word substitutions, insertions, and deletions. \methodtool{} follows the OLMoTrace procedure, but traces large batches offline and computes memorization metrics. Because the OLMoTrace code is not public\footnote{Only an infini-gram API wrapper is released: \url{https://github.com/allenai/infinigram-api}.}, we reproduced it and compared both tools on the same indexes and inputs (Appendix~\ref{sec:app-olmotrace}). They find the same longest matches and retrieve the source document of every verbatim text, and \methodtool{} is about 19$\times$ faster on verbatim text over the 52\,TB Dolma~3 index.

\section{Memorization Propensity Evaluation}

\subsection{Propensity and Capability Settings}
\label{sec:propensity-capability-settings}

Propensity evaluation of a behavior $b$ in a model $M$ compares two kinds of settings. In a \textit{propensity setting}, the inputs are not designed to elicit $b$. In a \textit{capability setting}, the inputs are designed to elicit $b$. Comparing the two tells us how much of the behavior a model can exhibit also appears without prompting.

For memorization, the capability setting is the prefix attack \citep{carlini2021extracting, karamolegkou2023copyright}: the model continues the beginning of a training document. The propensity settings use inputs that contain no training text: no prompt at all, a single common word, or short sentences that do not occur in the training corpus. We call these settings \textit{non-adversarial}. They are not a sample of real user queries, but controlled inputs that differ in how close they are to the training data (Section~\ref{sec:settings}).

\subsection{Propensity Metric}
\label{sec:propensity-metric}

A model's propensity toward a behavior should be read relative to its capability for it. Let $b$ be a behavior of a model $M$ and let $f_b \in [0,1]$ be a base metric that measures how much $b$ is exhibited, calculated over the prompts of one setting. Let $f_b^p(M)$ and $f_b^c(M)$ be the values of $f_b$ in a propensity setting $p$ and in a capability setting $c$.

Suppose two models have the same low $f_b^p$. If one of them has a high $f_b^c$, it clearly \textit{can} exhibit $b$ but does not do so without elicitation. That is stronger evidence of low propensity than for a model that shows little $b$ in both settings. We capture this with the propensity metric
\begin{equation}
\small
\label{eq:propensity-metric}
\PM_{f_b}(M) = \frac{f_b^p(M)}{f_b^p(M) + f_b^c(M)},
\end{equation}
and set $\PM_{f_b}(M)=0$ when $f_b^p(M)=f_b^c(M)=0$. $\PM_{f_b}$ is the share of the combined signal that is observed in the propensity setting. It lies in $[0,1]$:
\begin{itemize}
    \item $\PM_{f_b}=0$ when $f_b^p=0$: the behavior is never observed without elicitation, whatever the value of $f_b^c$.
    \item $\PM_{f_b}<0.5$ when $f_b^p<f_b^c$: the behavior is observed more under elicitation.
    \item $\PM_{f_b}=0.5$ when $f_b^p=f_b^c>0$: the behavior is observed equally in both settings (neutral point).
    \item $\PM_{f_b}>0.5$ when $f_b^p>f_b^c$, and $\PM_{f_b}=1$ when $f_b^p>0$ and $f_b^c=0$: the behavior appears without elicitation and would be missed by a capability-only test.
\end{itemize}
$\PM_{f_b}$ is unchanged when both inputs are multiplied by the same constant, so it can be compared across base metrics and corpora with different absolute levels. Appendix~\ref{sec:app-propensity-properties} gives its formal properties, its sensitivity to small changes in $f_b^p$ and $f_b^c$, and a comparison with simpler alternatives.
When using $\PM_{f_b}$ we also report bootstrap confidence intervals (Section~\ref{sec:experiment-metrics}). 
Any bounded base metric can be used, including verbatim and near-verbatim overlap or $n$-gram non-novelty \citep{merrill2024rustydawg}.

\paragraph{Propensity is not the amount of memorization.}
$\PM_{f_b}$ is a relative score. It does not say how often $b$ occurs: two models with the same $\PM_{f_b}$ can memorize very different amounts. We therefore always report $f_b^p$ and $f_b^c$ next to $\PM_{f_b}$, and the three values should be read together. Likewise, a low $\PM_{f_b}$ also does not show that a given document cannot be generated. A statement about the risk for a specific document or rightsholder would need the absolute values in \textbf{both} settings and an externally defined risk threshold.

\subsection{Tracing Generations to Training Data}
\label{sec:simpletrace}

\methodtool{} is our open-source batch implementation of the OLMoTrace span-matching procedure \citep{liu2025olmotrace} on top of infini-gram suffix-array indexes \citep{liu2024infini}. For a generation of $L$ tokens, \methodtool{} (1) finds every maximal span of the generation that occurs verbatim in the corpus, by looking up, at each token position, the longest matching span that starts there; (2) keeps the $\lceil 0.05L \rceil$ spans with the lowest unigram probability; (3) retrieves up to 10 training documents that contain each span; and (4) merges overlapping spans when their union also occurs in the corpus. It also looks up the whole generation, to detect full verbatim matches. Additional details are in Appendix~\ref{sec:app-simpletrace}. 
The check is deterministic: a span is reported only if it occurs in the corpus, and document sampling is seeded. On 360 validation queries built from documents of the three corpora, \methodtool{} returns the exact query text for every query and the source document for 357. In the other 3, the query occurs in more than 10 documents, all of which contain it (Appendix~\ref{sec:validation-methodtool}).

\section{Experimental Setup}
\label{sec:experimental-setup}

\subsection{Models, Training Corpora, and Indexes}
\label{sec:models-data}

\method{} needs models whose complete training data is public and can be indexed. This excludes most open-weight models. We use three such models from two independent lineages.

\textbf{Comma v0.1-2T} \citep{kandpal2026common} is a 7B model pre-trained for 2T tokens on the Comma v0.1 training dataset, drawn from Common Pile, a corpus of openly licensed English text. \textbf{DFM Decoder Open v0}\footnote{\url{https://huggingface.co/danish-foundation-models/dfm-decoder-open-v0-7b-pt}} continues the pre-training of Comma v0.1-2T for 30B tokens in three stages, with a fixed mixture of two-thirds Danish Dynaword \citep{enevoldsen2025dynaword} and one-third Common Pile (Appendix~\ref{sec:app-training-details}). We use the pair to study how continual pre-training on new data changes the memorization of the original data. \textbf{OLMo~3 32B} \citep{olmo2025olmo3} (\texttt{Olmo-3-1125-32B}) comes from a different developer, training pipeline, and corpus, and is more than four times larger. It is a base model pre-trained on Dolma~3.

We index the training corpora with infini-gram \citep{liu2024infini}, using the Llama-2 tokenizer; all span lengths in the paper are in Llama-2 tokens. The indexes cover the Comma v0.1 training dataset (463.6B Comma v0.1 tokens), Dynaword for DFM Decoder (6.83B Llama\,3 tokens), and the OLMo~3 pre-training data (about 6T Dolma~3 tokens, a 52\,TB index). Indexing Dolma~3 alone took about 60--70k CPU hours.
Appendix~\ref{sec:app-training-details} gives additional details.

\subsection{Evaluation Settings}
\label{sec:settings}

We use five settings for each model and training corpus. The first four are non-adversarial propensity settings; the last is the capability setting.

\textbf{Unconditional.} The model receives only its start-of-document token: \texttt{<|begin\_of\_text|>} for Comma and DFM Decoder, and \texttt{<|endoftext|>}, which separates documents in OLMo~3 training.

\textbf{Minimal cue.} The start-of-document token followed by one common word. For each generation, the word is drawn from the 100 most frequent words of the corpus language in wordfreq \citep{speer2022wordfreq}, weighted by frequency and capitalized (e.g., \textit{The}, \textit{Det}). The word only specifies the language.

\textbf{Generic.} Everyday sentences from Tatoeba \citep{tiedemann2020tatoeba}, a collaborative collection of sentences written by volunteers. We sample 2{,}000 unique English sentences (for Common Pile and Dolma~3) and 2{,}000 unique Danish sentences (for Dynaword). Filtering is described further down.

\textbf{Specific.} Sentences from sources that the tested model has not seen, chosen to be close to its training data. A source is used only if its content was created after the model's training data, or if the training-data documentation states that it was not included: Common Corpus collections \citep{langlais2025commoncorpus} outside Common Pile, Dynaword subsets not used for DFM Decoder, HPLT \citep{burchell2025hplt} and FinePDFs \citep{kydlicek2025finepdfs} for Dolma~3, and Wikipedia articles created after each model's data. We take one whole sentence per document, length-matched to the generic prompts, and keep within each source the sentences most similar to the training corpus (Appendix~\ref{sec:app-settings}).

\textbf{Prefix (capability).} Following the prefix attack \citep{carlini2021extracting, karamolegkou2023copyright}, we sample 1{,}000 training documents per corpus of at least 100 tokens and use their first 50 tokens as prompts.

\paragraph{No verbatim training text in non-adversarial prompts.}
To prevent a propensity setting from turning into a prefix attack, we remove every generic and specific prompt that occurs, as a whole, in the training index of the model. We look the whole sentence up in the index, under both tokenizations of its first word, and also check whether it is a substring of any document \methodtool{} retrieves for it. This removes 358 of the 2{,}000 English Tatoeba sentences for Common Pile, 826 for Dolma~3, and 336 of the 2{,}000 Danish sentences for Dynaword.

\paragraph{Specific prompts are closer to the training data.}
Measured as the cosine similarity of each prompt to its nearest of 10{,}000 random training documents (Qwen3-Embedding-8B; \citealp{zhang2025qwen3embedding}), specific prompts are closer to the training data than generic prompts for all three corpora (mean difference $+0.021$ to $+0.043$, all 95\% CIs above 0), also with bge-m3 \citep{chen2024bgem3}. They lie between generic prompts and real training sentences (Appendix~\ref{sec:prompt-validation}).

\paragraph{Generation.}
We sample 10 generations per prompt with vLLM \citep{kwon2023vllm} (temperature 0.7, top-$p$ 0.95, at most 256 new tokens; seed $42+g$ for generation $g$), which shows what a model tends to produce better than a single greedy output. Each prompted setting uses 1{,}000 prompts (all specific and prefix prompts, and 1{,}000 generic prompts drawn at random from the filtered sentences), which gives 10{,}000 generations; each prompt-free setting also has 10{,}000 generations. We trace only the continuation, which can be longer than the 50-token prefix.

\subsection{Metrics and Uncertainty}
\label{sec:experiment-metrics}

We trace every generation against its model's training index with \methodtool{} (up to 10 documents per span) and report three base metrics. The \textbf{full match ratio} (FMR) is the share of generations that occur in full in a training document \citep{carlini2023quantifying}. The \textbf{average longest span} (ALS) is the mean length, in tokens, of each generation's longest span in the corpus, 0 if there is none \citep{karamolegkou2023copyright}. The \textbf{near-verbatim recall} (NVR) is the mean adaptive nv-recall \citep{ahmed2026extracting} over all retrieved generation--document pairs (Appendix~\ref{sec:app-simpletrace-metrics}). We apply the propensity metric (Eq.~\ref{eq:propensity-metric}) to FMR and NVR, against the prefix setting of model and corpus.

We report 95\% percentile bootstrap intervals with 10{,}000 resamples. We resample prompts, each with all 10 of its generations, in prompted settings, and generations in prompt-free settings. For $\PM$, the two settings are resampled independently; runs that share prompts are compared with a paired bootstrap. A difference is significant when its 95\% interval excludes 0.

\paragraph{Degenerate outputs.}
We exclude generations that are loops of symbols, such as a `|' on every line, which match empty tables in every corpus without reproducing any content. A generation is excluded when its zlib compression ratio is below 0.1 \citep{carlini2021extracting} and less than half of its non-space characters are letters. This removes at most 75 of the 10{,}000 generations of a setting (Appendix~\ref{sec:app-long-matches}).

\section{Results}
\label{sec:results}

\begin{figure*}[t]
  \centering
  \includegraphics[width=0.85\textwidth]{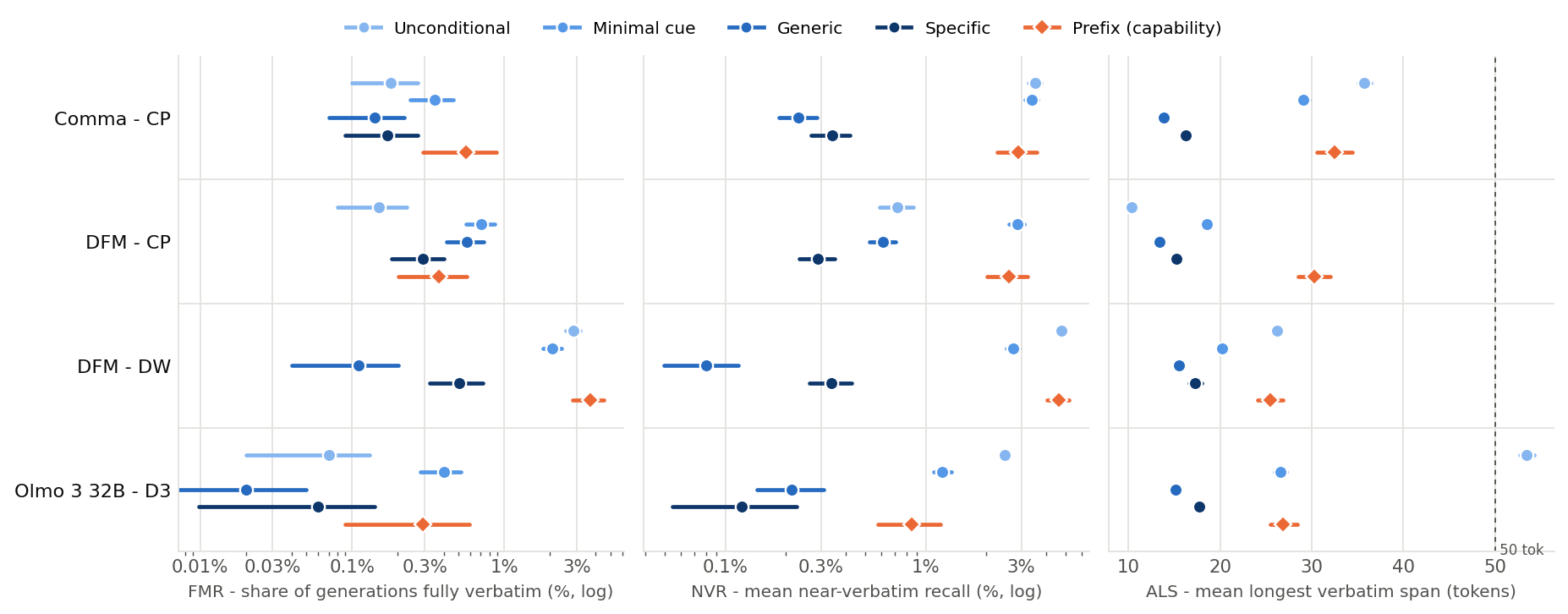}
  \caption{Memorization per model, training corpus, and setting, with 95\% bootstrap CIs. FMR: share of generations that occur in full in a training document; NVR: mean near-verbatim recall over retrieved documents (both on a log scale); ALS: mean longest span shared with the training data, in Llama-2 tokens. Blue: non-adversarial (propensity) settings, from the least (light) to the most (dark) targeted; orange diamonds: prefix attack (capability). Degenerate outputs are excluded. Values and significance tests: Table~\ref{tab:main-metrics}. CP: Common Pile, DW: Dynaword, D3: Dolma~3.}
  \label{fig:main-metrics}
\end{figure*}

\subsection{Prompt-Free Generation Reproduces as Much as Prefix Attacks; Natural Prompts Much Less}
\label{sec:res-absolute}

Figure~\ref{fig:main-metrics} shows NVR, FMR, and ALS for each combination of model, corpus, and setting. Table~\ref{tab:main-metrics} (Appendix~\ref{sec:app-main-tables}) shows the results with confidence intervals and marks significant differences from the prefix setting.

\paragraph{Natural-sentence prompts.}
Generic and specific prompts reproduce much less training text than prefix attacks. In all four model--corpus pairs, their NVR and ALS are significantly below the prefix setting: prefix NVR is 4 to 57 times higher, and the average longest span falls from 25--32 tokens under prefix attacks to 13--18 tokens. FMR is also lower, except for DFM Decoder on Common Pile, where the differences are not significant. Most of these full matches are short outputs, such as \textit{Thank you!}, that also occur in the training data. Counting only full matches of at least 50 tokens, generic and specific FMR is at most 0.0003, against 0.0014--0.0052 under prefix attacks (Section~\ref{sec:res-long}).

\paragraph{Prompt-free settings.}
Without a prompt, or with a single common word, the models reproduce about as much training text as under prefix attacks, and sometimes more. Unconditional NVR is at or above prefix NVR for Comma, DFM Decoder on Dynaword, and OLMo~3 (0.035 vs.\ 0.029, 0.048 vs.\ 0.046, and 0.025 vs.\ 0.009; only the last difference is significant), and OLMo~3's unconditional ALS is twice its prefix ALS (53.4 vs.\ 26.9 tokens). 
The remaining pair, DFM Decoder on Common Pile without a prompt, mostly writes Danish: the same unconditional generations contain a span of at least 50 tokens from Dynaword in 11\% of cases, but from Common Pile in only 1.9\%. With an English cue word, its NVR returns to the prefix level (0.029 vs.\ 0.026). Minimal-cue NVR does not differ significantly from prefix NVR in three of the four pairs.

\subsection{Propensity Depends on the Input}
\label{sec:res-propensity}

\begin{figure*}[!t]
    \centering
    \includegraphics[width=0.75\textwidth]{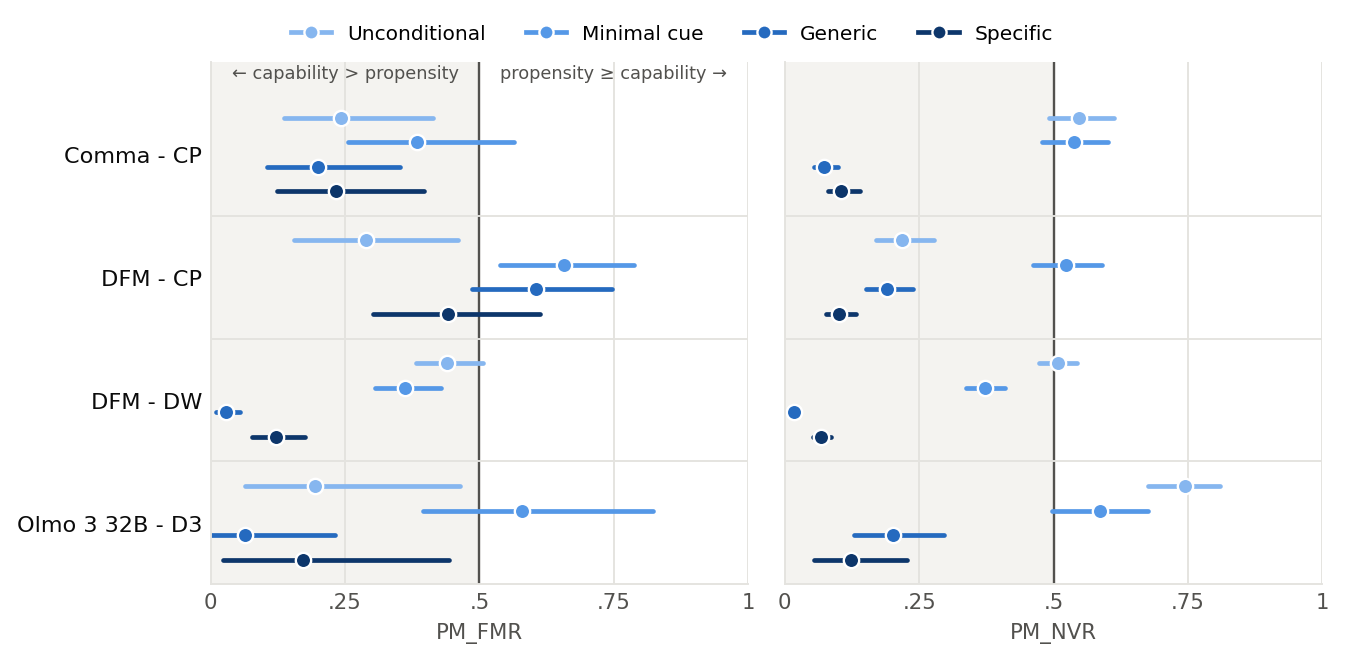} 
 \caption{Propensity scores of each non-adversarial setting against the prefix setting, for the full match ratio (FMR) and near-verbatim recall (NVR), with 95\% bootstrap intervals. Each point combines a propensity and a capability measurement. At 0.5 both settings reproduce equally; in the shaded half, the prefix attack reproduces more. CP: Common Pile, DW: Dynaword, D3: Dolma~3.}
    \label{fig:propensity-main-metrics}
\end{figure*}

Figure~\ref{fig:propensity-main-metrics} shows the propensity scores of each non-adversarial setting against the prefix setting; Table~\ref{tab:main-propensity} (Appendix~\ref{sec:app-main-tables}) gives the values.

\paragraph{Natural-sentence prompts.}
For generic and specific prompts, $\PM_{\mathrm{NVR}}$ ranges from 0.017 to 0.201, with every confidence interval below 0.5: the models show only a small share of the near-verbatim reproduction that prefix attacks reach. $\PM_{\mathrm{FMR}}$ is below 0.5 in seven of eight cases. The exception, DFM Decoder on Common Pile under generic prompts (0.606 [0.481,0.747]), comes from the short full matches described above; with full matches of at least 50 tokens, its score drops to 0 (Appendix~\ref{sec:app-long-matches}).

\paragraph{Prompt-free settings.}
Six of the eight prompt-free $\PM_{\mathrm{NVR}}$ scores are at or above the neutral value: Comma (0.548 unconditional, 0.538 minimal cue), DFM Decoder on Common Pile (0.523, minimal cue), DFM Decoder on Dynaword (0.508, unconditional), and OLMo~3 (0.744 and 0.586). The interval lies entirely above 0.5 only for OLMo~3 without a prompt. 
In these cases, a prefix attack on randomly chosen training documents does not bound what the model reproduces without adversarial prompting. Propensity is thus a property of model and input distribution together.

\subsection{Continual Pre-Training Reduces Long Matches Only Modestly}
\label{sec:res-continual}

DFM Decoder continues Comma's training on a mixture that keeps one third Common Pile. On the same prefix prompts, it reproduces Common Pile slightly less than Comma: NVR is lower by 0.0030 (CI [$-$0.0056, $-$0.0005]), ALS by 2.2 tokens (CI [$-$3.1, $-$1.4]), and 11.4\% instead of 13.6\% of generations contain a span of at least 50 tokens. FMR does not differ significantly, and both models reproduce the true continuation of the prompted document at similar rates (1.93\% vs.\ 2.15\% of generations). With an English cue word the reduction in long spans is larger (ALS $-$10.5 tokens; 6.5\% vs.\ 10.9\% of generations with a 50-token span). Under generic prompts, DFM Decoder has a higher NVR and FMR than Comma (+0.0038 and +0.0043), but these are short full matches: with a 50-token floor, generic FMR is 0.0000 for DFM Decoder and 0.0001 for Comma. The large unconditional drop (NVR $-$0.028) reflects the shift to Danish (Section~\ref{sec:res-absolute}). The continual pre-training setting considered thus reduces long verbatim reproduction of the original corpus only modestly. This is consistent with memorized examples being forgotten under further training \citep{jagielski2023measuring}, but less so when part of the original data stays in the mixture \citep{ibrahim2024simple}. Here the continuation is 30B tokens, 1.5\% of Comma's 2T. Across DFM Decoder's three training stages, Common Pile metrics stay mostly flat, while prefix memorization of Dynaword grows slightly with more Dynaword training (NVR 0.038 to 0.046; Appendix~\ref{sec:app-stages}).

\subsection{Long Matches: Length, Content, and Duplication}
\label{sec:res-long}

Short spans can match by chance: the prompt texts themselves, which are not in the training data, match spans of 6.5--8 tokens on average and at most 49 tokens. \citet{cooper2026extractable} show that OLMo~2 32B reproduces non-training 10-token suffixes about 24\% as often as training ones, but non-training 50-token suffixes almost never (0.02\% vs.\ 0.74\%). Following \citet{nasr2025scalable}, we therefore count long matches. R50 is the share of generations whose longest span has at least 50 tokens, not counting spans that loop one repeated unit (e.g., \textit{I'm sorry.} many times), and FMR50 counts only full matches of at least 50 tokens.

\paragraph{Long matches give the same picture.}
R50 is 0.4--0.9\% under generic prompts, 1.4--2.8\% under specific prompts, and 9.0--13.6\% under prefix attacks. Without a prompt it ranges from 1.9\% (DFM Decoder on Common Pile) to 41.4\% (OLMo~3). The corresponding $\PM_{\mathrm{R50}}$ is 0.035--0.220 for generic and specific prompts, and 0.565, 0.522, and 0.821 for unconditional generation of Comma, DFM Decoder on Dynaword, and OLMo~3. Specific prompts lead to more long matches than generic prompts in all four pairs (e.g., 2.8\% vs.\ 0.4\% on Dynaword), so closeness of the prompt to the training domain raises long verbatim reproduction, although it stays far below the prefix level.

\paragraph{What the long matches are.}
We inspected 100 matched generations per model and setting (full matches first, then the longest partial matches), categorized automatically by Claude Opus 5.5 and verified by us (Appendix~\ref{sec:app-inspection}). Without a prompt, long matches are mostly duplicated boilerplate: licenses and code (about half) and templated pages for Comma, coding exercises and code (72 of 100) for OLMo~3, and formulaic notices from 18th- and 19th-century newspapers for DFM Decoder on Dynaword. Specific prompts often lead into the boilerplate of their genre, such as an open-access license notice after an academic sentence. Generic prompts lead into Bible passages, reading-comprehension texts, or loops. Prior work also finds licenses and code among the most reproduced texts \citep{carlini2021extracting, aerni2024measuring}.

\paragraph{Duplication.}
The first 50 tokens of a long span usually occur many times in the training index (median 10--149 occurrences on Common Pile, 4--5 on Dynaword, and 18--71 on Dolma~3), in line with the role of duplication in memorization \citep{kandpal2022deduplicating, carlini2023quantifying}. Neither duplication nor code explains the prompt-free results, however. Counting only spans that occur once, unconditional R50 stays above prefix R50 for Comma (6.6\% vs.\ 5.1\%) and for DFM Decoder on Dynaword (4.6\% vs.\ 3.8\%). On natural-language generations only, prompt-free $\PM_{\mathrm{R50}}$ stays at or above 0.5 for Comma, DFM Decoder on Dynaword, and OLMo~3 (Appendix~\ref{sec:app-long-matches}).

\subsection{Repeated Sampling and Targeted Extraction}
\label{sec:res-sampling}

A single sample per prompt can underestimate extraction \citep{hayes2025measuring}. In at least one of their 10 generations, a span of at least 50 tokens appears for 3.3--7.2\% of generic, 10.7--16.2\% of specific, and 21.1--35.9\% of prefix prompts, the same ordering as per generation. Prefix generations reproduce the true 50-token continuation of their source document in only 1.0--2.2\% of cases, mostly for duplicated documents \citep{carlini2023quantifying}. Most long matches in the prefix setting therefore come from other documents, and tracing against the whole corpus counts reproduction that targeted extraction misses (Appendix~\ref{sec:app-long-matches}).

\section{Discussion}
Propensity depends on the input. Natural-sentence prompts, including sentences close to the training domains, rarely lead the evaluated models into long verbatim training text, and their propensity scores stay well below the neutral value. Generation without a prompt, or with a single common word, reaches the level of prefix attacks in three of four model--corpus pairs. Whether prefix attacks reproduce more than non-adversarial use therefore depends on the input, and the propensity metric shows this for each model, corpus, and setting.

What the models reproduce without a prompt is mostly duplicated boilerplate: licenses, code, templates, and formulaic notices. This content is less sensitive than rare text, but for Comma and for DFM Decoder on Dynaword the prompt-free settings also reproduce spans that occur only once in the corpus, at rates similar to prefix attacks. Low propensity under one input distribution is therefore no guarantee for another.

Metric choice is also relevant. FMR counts full matches of any length, so under natural prompts it partly reflects short outputs, and short spans can match by chance. Complementing NVR and FMR with long-match metrics, such as R50 and FMR50 (Section~\ref{sec:res-long}), and reporting the absolute values in both settings next to each propensity score, can make such comparisons easier to interpret.

\section{Conclusion}
We introduced \method{}, a framework that evaluates memorization in paired settings: the same model and training corpus are tested with non-adversarial prompts and with prefix attacks, and every generation is checked against the full training corpus with \methodtool{}. We evaluated three fully open base models (Comma and DFM Decoder, 7B, and OLMo~3, 32B) against their indexed training corpora, from 6.8B tokens (Dynaword) to about 6T tokens (Dolma~3, a 52\,TB index), with five settings and 10{,}000 generations per model, corpus, and setting (200{,}000 traced generations in total). Natural-sentence prompts reproduce much less training text than prefix attacks (propensity scores of 0.02--0.20 for near-verbatim recall). Prompt-free generation reproduces about as much as prefix attacks in three of four model--corpus pairs, mostly duplicated boilerplate. The continual pre-training setting considered reduces long verbatim reproduction of the original corpus only modestly. These results support reporting propensity and capability together, since neither alone describes what a model reproduces.

\section{Limitations}
\method{} needs the full training corpus of a model, indexed for search, so it cannot be applied to models whose training data is not released. The propensity metric itself does not depend on corpus search and could be combined with logit-, weight-, or probability-based memorization measures, but we have not tested this.

Our models cover two lineages: Comma with its continual pre-training DFM Decoder, and OLMo~3. Like many works on training-data memorization \citep{carlini2021extracting, carlini2023quantifying, huang2024demystifying, hayes2025measuring}, we focus on base models only. In this context, a post-trained model would not be comparable with the others: its post-training data would also have to be indexed, and its matches with the pre-training data would mix the effects of pre-training and post-training. Studying this properly would need its own experimental design, which we leave to future work. Post-training changes how a model responds to prompts and may change its propensity to reproduce training data in either direction, so our results should not be transferred to post-trained models without testing them.

Our non-adversarial settings are controlled inputs, not a sample of real user prompts. Real use includes prompts between our specific and prefix settings, for example prompts that mention a document's topic, title, or author without quoting it; measuring them is future work.

\methodtool{} detects verbatim spans and word-level near-verbatim overlap. It does not detect paraphrases or other non-literal reproduction. Our long-match threshold of 50 tokens is motivated by prior work on targeted extraction, and is not calibrated with matched non-training text in our setting. The content analysis covers the longest matches of each setting, categorized automatically and verified by the authors, and does not describe all generations.

We focus on verbatim and near-verbatim reproduction because these are the most harmful cases of training-data leakage, and because the established memorization metrics and verbatim-search methods we build on measure them. Extending the framework to softened metrics is left as future work.

\section{Ethical Considerations}
We evaluate models whose training data is public and released for research. Comma and DFM Decoder are trained on openly licensed data. OLMo~3 is trained on Dolma~3, a mostly web-crawled corpus released under the ODC-By license; this license covers the collection, and unlike Common Pile, not every document in it is openly licensed. Our results show that prefix attacks can surface memorized content even when non-adversarial reproduction is low, so capability tests remain necessary alongside propensity tests. We release \methodtool{} as open source to support reproducible research. Tracing outputs to training data could in principle be misused, but it requires full access to the training corpus, and we believe the accountability benefits outweigh this risk.

\method{} is a research measurement, not a compliance or audit instrument. Its scores do not correspond to legal thresholds under the GDPR, the EU AI Act, or copyright law, and a low propensity score is not evidence that a given document or rightsholder is safe. Turning propensity measurements into an audit standard would require legal and policy expertise and validation that are beyond this paper; we leave this interdisciplinary work to future research. Low memorization propensity should not be used to downplay potential copyright infringement.

\section*{Acknowledgements}
The research was supported in part by the Danish Foundation Models project, funded by the Danish government.
This research was further supported in part by the MIST project, funded by the Novo Nordisk Foundation under grant reference number NNF25OC0103204. 
Part of the computation done for this project was
performed on the UCloud interactive HPC system
managed by the eScience Center at the
University of Southern Denmark.

\bibliography{custom}

\appendix

\section{Extended Related Work}
\label{sec:app-extended-related-work}

\paragraph{Memorization.}
Beyond the split into detection and extraction (Section~\ref{sec:related-work}), memorization studies differ in the models they target and in how they measure memorization. Target models range from closed or commercial models \citep{ahmed2026extracting} to open models \citep{carlini2021extracting, panda2025privacy, cooper2025extracting}. Measurement methods range from model-internal signals, such as activations, weights, or output probabilities \citep{huang2024demystifying, shi2024mink, zhang2024divergence, menta2025analyzing}, to comparisons with external texts, such as books or training data \citep{kassem2025alpaca, kandpal2022deduplicating, kiyomaru2024comprehensive}. Detection methods based on token probabilities \citep{shi2024mink, zhang2024divergence} usually set their decision threshold on known members and non-members. \method{} belongs to the second group: it compares generations with the full training corpus. Its propensity metric could also be applied to model-internal measures, which we have not tested.

\paragraph{Reproduction without adversarial prompting.}
\citet{aerni2024measuring} measure the share of a generation's characters that fall inside a substring of at least 50 characters found in a 10\,TB web snapshot. Under benign prompts, up to 15\% of the output of chat models overlaps with this snapshot, and code is prominent among the longest matches. Because the snapshot stands in for the unknown training data, these rates are lower bounds. \citet{nasr2025scalable} prompt open models with random five-token Wikipedia snippets and find that 0.16--1.4\% of the generated tokens belong to spans of at least 50 tokens that occur in the training data. \citet{merrill2024rustydawg} find that, for $n>4$, text generated by language models contains fewer novel $n$-grams than human-written text. \method{} differs from these studies in two ways. It checks generations against the actual training corpus of each model, and it evaluates the same model and corpus under a capability setting, so the two can be compared directly.

\paragraph{Propensity vs.\ capability.}
Capability tests may not predict deployment behavior (Section~\ref{sec:related-work}) for three reasons studied so far. Models can hide or fail to reveal latent capabilities under elicitation \citep{greenblatt2024stress, hofstatter2025elicitation}. They can strategically underperform on evaluations \citep{vanderweij2024sandbagging}. And they can adapt their behavior when they detect that they are being evaluated \citep{needham2025large}. In agentic safety, propensity evaluations separate whether models are merely capable of scheming or misaligned behavior from whether they are likely to show it under deployment-like prompts, goals, tools, and oversight conditions \citep{meinke2024frontier, hopman2026evaluating, naik2025agentmisalignment, jarviniemi2026propensity}. Memorization differs from these behaviors in one respect: with an indexed training corpus, every instance of reproduction can be verified directly, so propensity can be measured without a judge or a behavioral rubric.

\section{Additional Memorization Metrics}
\label{sec:app-additional-memorization-metrics}
\textbf{$k$-Eidetic memorization} \citep{carlini2021extracting} defines a string $s$ as memorized if it is extractable from $f_\theta$ and occurs in at most $k$ training examples: $|\{x \in \mathcal{X} : s \subseteq x\}| \leq k$.
\textbf{Near-duplication count} \citep{kiyomaru2024comprehensive} counts training documents whose token-frequency vectors satisfy weighted Jaccard similarity $J_W(a,b) \geq 0.6$ with a generated span. \textbf{ROUGE-L} \citep{kassem2025alpaca} and \textbf{Token Accuracy} \citep{menta2025analyzing}, the fraction of suffix tokens matching the greedy continuation, round out the set by capturing partial reproduction at different granularities.

\section{Comparison with OLMoTrace}
\label{sec:app-olmotrace}
\paragraph{Setup.} The OLMoTrace code is not public, so we reimplemented it from its paper \citep{liu2025olmotrace}, including the BM25 ranking of retrieved documents. Both tools ran on the same three indexes, with the same tokenizer, unigram tables, and number of retrieved documents per span (10), and with the same parallelism (8 processes with 4 threads each, on 32 CPUs and 96\,GB of memory). The inputs were 64 pieces of verbatim training text per corpus (the first 256 tokens of sampled documents), 300 model generations traced against Dynaword, and 100 short English sentences traced against Common Pile and Dolma~3. \methodtool{} ran in \texttt{mixed} mode, as in our experiments. Table~\ref{tab:olmotrace} summarizes the results.

\paragraph{Results.} The longest-match lookup, the core step of both tools, returned the same correct length on all 19{,}080 suffixes we tested. Both tools retrieved the source document of all 192 verbatim texts. \methodtool{} also flagged every input whose whole text occurs in the index (216 inputs, none missed). On the model generations, both tools order the prompt settings in the same way, with generic and specific prompts below prefix prompts, and they rank generations by the share of matched tokens in a similar order (Spearman 0.85). The highlighted spans differ mainly because \methodtool{} lets a span continue across a sentence end, which OLMoTrace does not; this explains 88\% of the tokens that only \methodtool{} highlights. On short sentences the two tools highlight almost the same tokens. On verbatim text over the 65-shard Dolma~3 index, OLMoTrace makes about five times more index queries and its memory use grows without bound: it took 141\,s for 64 texts against 7.4\,s for \methodtool{}, and with 32 processes it ran out of memory.

\begin{table*}[t]
\centering
\small
\setlength{\tabcolsep}{4pt}
\begin{tabular}{@{}lrr@{}}
\toprule
& \textbf{OLMoTrace} & \textbf{\methodtool{}} \\
\midrule
Correct longest match (19{,}080 suffixes) & 100\% & 100\% \\
Source document found (192 texts) & 192 & 192 \\
Whole-text matches flagged (216) & -- & 216 \\
Token Jaccard vs.\ OLMoTrace, short sentences & -- & 0.98--0.99 \\
Token Jaccard vs.\ OLMoTrace, verbatim text & -- & 0.81--0.97 \\
Token Jaccard vs.\ OLMoTrace, generations & -- & 0.75 \\
Dolma~3, 64 verbatim texts: time (s) & 141.3 & 7.4 \\
Dolma~3, 64 verbatim texts: page tables (GB) & 78.8 & 7.6 \\
\bottomrule
\end{tabular}
\caption{OLMoTrace reimplementation vs.\ \methodtool{} (\texttt{mixed} mode) on the same indexes and inputs. Times use a warm cache.}
\label{tab:olmotrace}
\end{table*}

\section{Properties of the Propensity Metric}
\label{sec:app-propensity-properties}
We write $f_p=f_b^p(M)$ and $f_c=f_b^c(M)$ for the base metric in the propensity and capability settings, with $f_p,f_c\in[0,1]$. The propensity metric is
\begin{equation}
\label{eq:app-pm}
\begin{split}
    \PM=\begin{cases}\frac{f_p}{f_p+f_c} & f_p+f_c>0\\ 0 & f_p=f_c=0\end{cases}
\end{split}
\end{equation}

\paragraph{Proposition.} $\PM$ has the following properties.
\begin{itemize}
    \item \textbf{Bounded.} $0\le f_p\le f_p+f_c$, so $\PM\in[0,1]$. The convention for $f_p=f_c=0$ keeps it in $[0,1]$.
    \item \textbf{Monotone.} For $f_p+f_c>0$,
    \begin{equation}
    \begin{split}
    \frac{\partial \PM}{\partial f_p}=\frac{f_c}{(f_p+f_c)^2}\ge 0,
    \\
    \frac{\partial \PM}{\partial f_c}=-\frac{f_p}{(f_p+f_c)^2}\le 0 .
    \end{split}
    \end{equation}
    $\PM$ grows when more behavior is observed without elicitation, and shrinks when more is observed under elicitation.
    \item \textbf{Scale invariant.} For any $k>0$, $\frac{kf_p}{kf_p+kf_c}=\frac{f_p}{f_p+f_c}$. Rescaling both measurements by the same factor, for example by a stricter matching criterion, does not change $\PM$. This allows comparisons across base metrics and corpora with different absolute levels.
    \item \textbf{Neutral point.} For $f_p+f_c>0$, $\PM=0.5$ if and only if $f_p=f_c$.
\end{itemize}

\paragraph{Endpoints.} $\PM=0$ whenever $f_p=0$. This covers two different cases: the behavior is observed in neither setting ($f_c=0$), or only under elicitation ($f_c>0$). $\PM=1$ when $f_p>0$ and $f_c=0$: the behavior appears without elicitation but not under the capability test, so a capability-only evaluation would miss it. The endpoints describe the relation between the two settings, not overall risk. Cases with the same $\PM$ are told apart by the reported $f_p$ and $f_c$.

\paragraph{Relation to the ratio.} For $f_c>0$, let $r=f_p/f_c$. Then
\begin{equation}
\PM=\frac{r}{1+r} \qquad \log\frac{\PM}{1-\PM}=\log f_p-\log f_c .
\end{equation}
$\PM$ is therefore a bounded, increasing function of the ratio $r$: it orders model--corpus pairs exactly as $r$ does. Unlike $r$, it is bounded and defined when $f_c=0$. The raw difference $f_p-f_c$ can order pairs differently, because it depends on the absolute level of the base metric: the same relative gap gives a larger difference when both values are large.

\paragraph{Sensitivity.} To first order, a small change $(\Delta f_p,\Delta f_c)$ changes $\PM$ by
\begin{equation}
\Delta \PM \approx \frac{f_c\,\Delta f_p - f_p\,\Delta f_c}{(f_p+f_c)^2}.
\end{equation}
The change is inversely proportional to $(f_p+f_c)$. When both base values are small, small absolute changes move $\PM$ a lot. For example, with $f_p=0.002$ and $f_c=0.004$, $\PM=0.33$, and increasing $f_p$ by $0.001$ gives $\PM=0.43$. With $f_p=0.02$ and $f_c=0.04$, the same increase gives $0.34$. This is why we report bootstrap intervals for $\PM$ (Section~\ref{sec:experiment-metrics}): rare events such as full matches give wide intervals, and differences in $\PM$ should be judged against these intervals.

\paragraph{Alternatives.} The difference $f_p-f_c$ lies in $[-1,1]$ and depends on the absolute scale of the base metric. The ratio $f_p/f_c$ is undefined for $f_c=0$ and unbounded. $\PM$ keeps the ordering of the ratio while being bounded, defined for $f_c=0$, and centered at $0.5$. Other transformations with the same properties exist; we present $\PM$ as one simple choice, not the only one.

\paragraph{Comparisons across corpora.}
The propensity metric compares corpora with different absolute levels. Under generic prompts, DFM Decoder has a raw NVR of 0.0061 on Common Pile and 0.0008 on Dynaword, both small. Relative to what prefix attacks reach on the same corpus (0.026 and 0.046), the scores differ by an order of magnitude: $\PM_{\mathrm{NVR}}=0.190$ on Common Pile, against 0.017 on Dynaword.

\section{\methodtool{} Details}
\label{sec:app-simpletrace}
\methodtool{} traces each generation against an infini-gram index \citep{liu2024infini}. Generations and indexes use the Llama-2 tokenizer. The steps follow OLMoTrace \citep{liu2025olmotrace}; we note where they differ (see also Appendix~\ref{sec:app-olmotrace}).

\paragraph{Step 1: maximal matching spans.}
\methodtool{} finds every maximal span of the generation that occurs verbatim in the corpus. For each token position of a generation with $L$ tokens, it finds the longest span starting there that occurs in the corpus. Any matching span inside the generation is contained in one of these, so together they cover all matches. One query on the whole remaining text is enough: if it does not occur, the longest matching prefix is shared with one of the two suffix-array entries around the position where it would be inserted, so only these two entries per shard are checked. The match found at one position also gives a lower bound for the next position, which often makes the full lookup unnecessary; the results are identical. Spans that lie inside another span are removed. \methodtool{} has two modes. In \texttt{text} mode, a span must start and end at word boundaries and is cut at the first sentence boundary. In \texttt{mixed} mode, which we use in all experiments, these rules are dropped so that code, markup, and lists are not split, and spans must be at least 4 tokens long.

\paragraph{Step 2: rare spans.}
Each span is scored by the product of the unigram probabilities of its tokens, precomputed over the corpus. The $K=\lceil 0.05L\rceil$ rarest spans are kept, which removes common phrases that match many documents by chance.

\paragraph{Step 3: document retrieval.}
For each kept span, \methodtool{} retrieves up to 10 enclosing documents, as a window of up to $5L$ tokens around the match. If the span occurs more often, the 10 occurrences are sampled with a random generator seeded by a fixed seed and the generation text, so results do not depend on run order. Each document is labeled as a full match (it contains the whole generation), a full match after light whitespace and Unicode normalization, or a partial match.

\paragraph{Step 4: merging.}
Overlapping spans are merged only if their union also occurs in the corpus, and documents are then retrieved again for the union. OLMoTrace merges all overlapping spans, which can create a span that never occurs in the corpus as a whole.

\paragraph{Full matches.}
In \texttt{mixed} mode, \methodtool{} also looks up the whole generation. If it occurs, its documents are attached as a full-generation span. Otherwise, the longest and rarest spans are used as anchors to search for documents that contain the generation after normalization.

\paragraph{Batch processing.}
Generations are distributed over worker processes, and each worker runs the queries of one generation on several threads. On very large memory-mapped indexes, the operating system's page tables grow with every lookup. Each worker therefore reopens its index once its page tables exceed 1\,GB, which keeps memory bounded without changing results. \methodtool{} also aggregates the per-document results into the metrics of Appendix~\ref{sec:app-simpletrace-metrics}. Compared with earlier tracing scripts \citep{wolfe2025olmotracegist}, it adds index building, unigram precomputation, batch processing, and metric aggregation.

\section{Validation of \methodtool{}}
\label{sec:validation-methodtool}
We validate \methodtool{} at two levels. First, 25 unit tests on a small corpus with known document identifiers check exact span recovery from the start, middle, and end of documents, matches across several documents, queries with no match, full-document matches, and the summary statistics.

Second, we validate on the three real indexes. We sample documents from each index (20 from Common Pile, 20 from Dolma~3, 50 from Dynaword; seed 42) and build four queries per document: the full document (up to 1{,}000 tokens) and three 128-token windows at its start, middle, and end. This gives 360 queries. A query \textit{passes} if \methodtool{} retrieves the source document, or if every retrieved document contains the query text exactly. The second case covers duplicated text that occurs in more than the 10 documents retrieved per span. Table~\ref{tab:validation} shows the results. All 360 queries pass. The source document is retrieved for 357 queries. In the 3 Dolma~3 cases where it is not, the query occurs in more than 10 documents and all retrieved documents contain it.

\begin{table}[t]
\centering
\small
\begin{tabular}{@{}lrrrr@{}}
\toprule
& & \multicolumn{2}{c}{\textbf{Source doc.\ retrieved}} & \\
\cmidrule(lr){3-4}
\textbf{Corpus} & \textbf{Queries} & \textbf{Full} & \textbf{Partial} & \textbf{Pass} \\
\midrule
Common Pile & 80 & 1.00 & 1.00 & 1.00 \\
Dolma~3 & 80 & 1.00 & 0.95 & 1.00 \\
Dynaword & 200 & 1.00 & 1.00 & 1.00 \\
\bottomrule
\end{tabular}
\caption{End-to-end validation of \methodtool{}. Each sampled document gives one full and three partial (start, middle, end) queries. Dolma~3 partial queries: start 1.00, middle 0.90, end 0.95.}
\label{tab:validation}
\end{table}

\section{Training and Indexing Details}
\label{sec:app-training-details}
\paragraph{DFM Decoder.} Stage~1 used a batch size of 262\,144 tokens for 37\,852 steps ($\text{lr}=1\text{e}{-5}$, constant); Stages~2 and~3 used 524\,288 tokens over 18\,926 steps each, and Stage~3 applied square-root decay from $1\text{e}{-5}$. The released checkpoint is Stage~3.

\paragraph{Indexes.} \textbf{Common Pile}: the Comma v0.1 training dataset (521\,GB, 463.6B Comma v0.1 tokens), indexed in three parts with 128 CPUs and 350\,GB of memory (about 3 hours per part). \textbf{Dynaword}: the version used to train DFM Decoder (5.7M documents, 6.8B Llama\,3 tokens, 10.5\,GB), indexed with 16 CPUs and 84\,GB of memory in about 3 hours. \textbf{Dolma~3}: the OLMo~3 pre-training data (about 6T Dolma~3 tokens), indexed in 26 parts that are queried together as one 65-shard index (52\,TB).

\section{Construction of the Specific Prompt Sets}
\label{sec:app-settings}
\paragraph{Eligible sources.} A source is used for a model only if (a) its content was created or published after the model's training data (a crawl date does not count, since a recent crawl can repeat an older page), or (b) the documentation of the model or its training data states that the source was not included. The rule covers all training data of each model. Table~\ref{tab:specific-sources} lists the sources.

\begin{table*}[t]
\centering
\small
\begin{tabular}{@{}lp{0.62\textwidth}c@{}}
\toprule
\textbf{Corpus} & \textbf{Sources} & \textbf{Rule} \\
\midrule
Dynaword & kb\_administrative\_publication, kb\_historical\_letters, municipality\_meetings, hvadvilduhelst, tidsskrift-dk, dakultur, mosel\_voxpopuli, mosel\_youtubecommons, folketingets-dokumenter, kalliope, logir: added to Dynaword after v1.2.12 & (b) \\
& depbank, jvj, nordjyllandnews, synne: in v1.2.12 but excluded from DFM Decoder training & (b) \\
& Danish Wikipedia articles created after 2025-09-19 & (a) \\
\midrule
Common Pile & Common Corpus collections SEC, WTO, GATT\_library, Eurlex, Eurovoc, TEDEUTenders, VoxPopuli, dotgov, US-PD-Newspapers, NewZealand-PD-Newspapers, whose origin is none of the Common Pile sources & (b) \\
& English Wikipedia articles created after 2025-06-06 & (a) \\
\midrule
Dolma~3 & HPLT (top quality bins) and FinePDFs (English), not in the OLMo~3 training data & (b) \\
& English Wikipedia articles created after 2025-11-20 & (a) \\
\bottomrule
\end{tabular}
\caption{Sources of the specific prompt sets. Rule (a): content created after the training data; (b): documented as not used for training.}
\label{tab:specific-sources}
\end{table*}

\paragraph{Filters.} From each source we draw up to 4{,}000 random documents (8{,}000 for HPLT and FinePDFs). English documents need a FineWeb-Edu quality score of at least 2 \citep{penedo2024fineweb}, which keeps roughly the top 30--45\% of each source; the Danish sources are curated and not scored. From each document we take one whole sentence whose length is within one token (10\% above 10 tokens) of a length drawn from the generic set, so both sets have similar lengths. Sentences must be in the corpus language, judged by a function-word share and by an XLM-R language classifier,\footnote{\url{https://huggingface.co/qanastek/51-languages-classifier}} and must not occur verbatim in the training index (Section~\ref{sec:settings}).

\paragraph{Similarity selection.} We embed each remaining candidate and 10{,}000 training documents (sampled with seed 43) with Qwen3-Embedding-8B. Within each source, we keep the candidates with the highest similarity to their nearest training document, after subtracting the mean similarity of candidates of the same length, so that the selection does not favor longer sentences. All sources of a corpus get equal shares; sources with too few candidates contribute all of them, and their unused share goes to the other sources. Before selection, the Common Pile and Dolma~3 candidates were no closer to the training data than Tatoeba (mean nearest-document similarity 0.460 and 0.459, against 0.461 and 0.463). The eligible sources are, by construction, ones the training data does not contain, which is why the selection step is needed. The evaluation in Appendix~\ref{sec:prompt-validation} uses an independent sample of training documents (seed 42).

\section{Similarity of the Prompt Sets to the Training Data}
\label{sec:prompt-validation}
Table~\ref{tab:prompt-similarity} compares the prompt sets with the training data. For each corpus, we sample 10{,}000 random training documents and embed them, together with every prompt, using two embedding models. A prompt's similarity is the cosine similarity to its nearest document. As a reference for in-domain text, we also embed 2{,}000 sentences drawn from other training documents of the same corpus. With both models and for all three corpora, specific prompts are closer to the training data than generic prompts. They lie between generic prompts and in-domain training sentences. None of the generic or specific prompts occurs verbatim in the training data (Section~\ref{sec:settings}). With Qwen3-Embedding-8B, the mean difference between specific and generic prompts is $+0.022$ [95\% CI $+0.018$, $+0.027$] for Common Pile, $+0.021$ [$+0.017$, $+0.025$] for Dolma~3, and $+0.043$ [$+0.039$, $+0.048$] for Dynaword.

\begin{table}[t]
\centering
\small
\setlength{\tabcolsep}{4pt}
\begin{tabular}{@{}llrrr@{}}
\toprule
\textbf{Corpus} & \textbf{Set} & \textbf{Tokens} & \textbf{Qwen3} & \textbf{bge-m3} \\
\midrule
\multirow{3}{*}{CP} & Generic & 11.6 & 0.463 & 0.544 \\
 & Specific & 12.9 & 0.485 & 0.553 \\
 & Training & -- & 0.517 & 0.572 \\
\midrule
\multirow{3}{*}{D3} & Generic & 11.4 & 0.464 & 0.532 \\
 & Specific & 12.8 & 0.485 & 0.556 \\
 & Training & -- & 0.489 & 0.563 \\
\midrule
\multirow{3}{*}{DW} & Generic & 13.2 & 0.455 & 0.527 \\
 & Specific & 15.3 & 0.499 & 0.540 \\
 & Training & -- & 0.573 & 0.607 \\
\bottomrule
\end{tabular}
\caption{Mean cosine similarity of each prompt to its nearest of 10{,}000 random training documents (1{,}000 prompts per set). Training: 2{,}000 sentences from other training documents of the same corpus (in-domain reference). Tokens: mean prompt length in Llama-2 tokens.}
\label{tab:prompt-similarity}
\end{table}

\section{\methodtool{} Metrics}
\label{sec:app-simpletrace-metrics}
\methodtool{} writes, for every generation, its final spans and the retrieved documents, and a summary over all generations. The paper uses three summary metrics.

\textbf{FMR} is the fraction of generations for which at least one retrieved document contains the full generation verbatim (without normalization).

\textbf{ALS} is the mean, over generations, of the length in Llama-2 tokens of the longest final span; generations without spans count as 0.

\textbf{NVR} averages, over all retrieved documents of all generations, the adaptive near-verbatim recall between the generation and the document. Following \citet{ahmed2026extracting}, we find the word blocks shared by the generation and the document, merge blocks separated by small gaps, drop short blocks, and repeat the merge and drop with larger thresholds. The recall is the number of generation words in the kept blocks divided by the number of words in the generation. Because our generations and documents have very different lengths, the thresholds scale with the number $n$ of words in the generation, clipped to a fixed range: minimum block length $\ell_1=\mathrm{clip}(\mathrm{round}(0.15n),2,20)$ and $\ell_2=\mathrm{clip}(\mathrm{round}(0.35n),4,100)$; maximum gap $\tau^{\mathrm{gap}}_1=\mathrm{clip}(\mathrm{round}(0.02n),1,6)$ and $\tau^{\mathrm{gap}}_2=\mathrm{clip}(\mathrm{round}(0.08n),2,12)$; maximum alignment offset $\tau^{\mathrm{align}}_1=\mathrm{clip}(\mathrm{round}(0.01n),1,3)$ and $\tau^{\mathrm{align}}_2=\mathrm{clip}(\mathrm{round}(0.03n),1,5)$.

\textbf{R50} is the share of generations whose longest span in the training data has at least 50 Llama-2 tokens, not counting repetition loops (Section~\ref{sec:res-long}); \textbf{FMR20} and \textbf{FMR50} count only full matches of at least 20 and 50 tokens.

The summary also contains span-length histograms, counts of unique and fully matching documents, and optional $k$-eidetic rates \citep{carlini2021extracting}. The repository documents all fields.

\section{Long Matches, Duplication, and Sampling: Details}
\label{sec:app-long-matches}

\paragraph{Degenerate outputs.} Table~\ref{tab:app-degenerate} lists the generations excluded by the degenerate-output filter (Section~\ref{sec:experiment-metrics}).

\begin{table}[t]
\centering
\scriptsize
\setlength{\tabcolsep}{2pt}
\begin{tabular}{@{}llccccc@{}}
\toprule
\textbf{Model} & \textbf{Data} & \textbf{Uncond.} & \textbf{Min.\ cue} & \textbf{Generic} & \textbf{Specific} & \textbf{Prefix} \\
\midrule
Comma & CP & 13 (1) & 1 (0) & 3 (0) & 5 (2) & 12 (0) \\
\midrule
DFM & CP & 1 (0) & 3 (0) & 3 (2) & 75 (38) & 18 (0) \\
\midrule
DFM & DW & 1 (0) & 1 (0) & 16 (8) & 74 (34) & 4 (0) \\
\midrule
OLMo~3 & D3 & 8 (0) & 0 & 0 & 0 & 12 (0) \\
\midrule
DFM Stage 1 & CP & 9 (5) & 2 (0) & 12 (5) & 68 (40) & 21 (0) \\
\midrule
DFM Stage 2 & CP & 4 (0) & 1 (0) & 5 (2) & 60 (28) & 15 (0) \\
\midrule
DFM Stage 1 & DW & 9 (4) & 2 (0) & 13 (6) & 68 (28) & 4 (0) \\
\midrule
DFM Stage 2 & DW & 4 (0) & 1 (0) & 16 (3) & 51 (16) & 3 (0) \\
\bottomrule
\end{tabular}
\caption{Degenerate generations excluded per run, out of 10{,}000 each; full matches among them in parentheses.}
\label{tab:app-degenerate}
\end{table}

\paragraph{Chance level.} Table~\ref{tab:app-floor} gives the matches of the prompt texts themselves, which do not occur in the training data.

\begin{table}[t]
\centering
\scriptsize
\setlength{\tabcolsep}{2pt}
\begin{tabular}{@{}lcccc@{}}
\toprule
\textbf{Index} & \textbf{Generic ALS ($n$)} & \textbf{Specific ALS ($n$)} & \textbf{$\geq$20 tok.} & \textbf{Max} \\
\midrule
Common Pile & 6.59 (1642) & 6.54 (1000) & 0.0\% & 19 \\
Dolma 3 & 7.35 (1174) & 7.32 (1000) & 0.8\% & 49 \\
Dynaword & 7.60 (1664) & 8.03 (1000) & 0.8\% & 30 \\
\bottomrule
\end{tabular}
\caption{Matches of the prompt texts themselves, which do not occur in the training data: average longest span of generic and specific prompts, share of specific prompts with a $\geq$20-token span, and the longest span of any specific prompt.}
\label{tab:app-floor}
\end{table}

\begin{figure*}[t]
  \centering
  \includegraphics[width=0.8\textwidth]{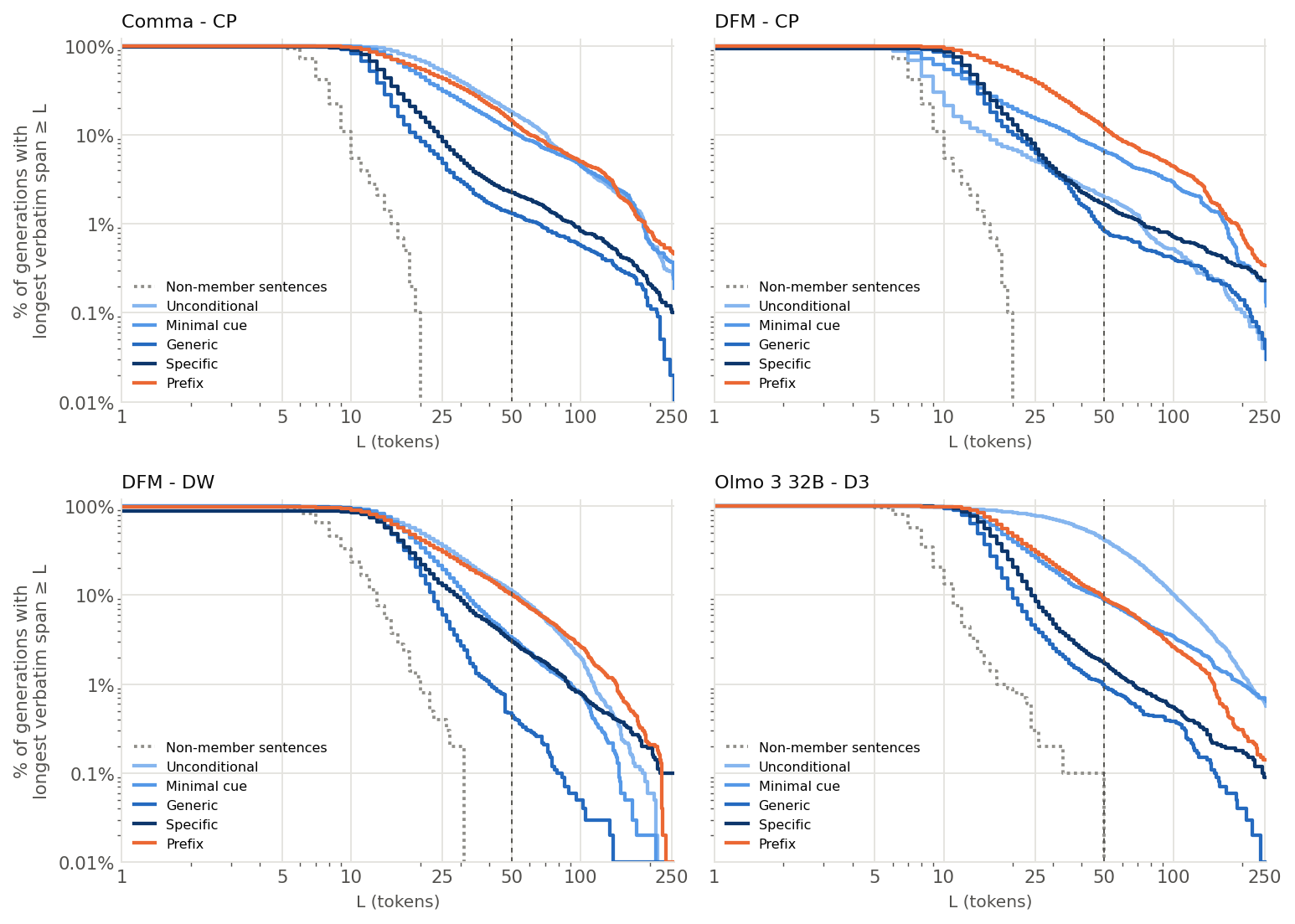}
  \caption{Share of generations whose longest span shared with the training data has at least $L$ tokens, per setting (log--log). The area under each curve equals ALS. The value at $L=50$ (dashed line) is close to R50, which in addition excludes repetition loops. Grey dotted: the specific prompt texts themselves, which do not occur in the training data, as a chance level for short matches.}
  \label{fig:app-survival}
\end{figure*}

\paragraph{Long matches and full matches by length.} 
Figure~\ref{fig:app-survival} shows the full distribution of longest-span lengths.
Tables~\ref{tab:app-long}, \ref{tab:app-fmr-length}, and~\ref{tab:app-pm-long} report R20, R50, NVR50 (NVR counting only documents retrieved through spans of at least 50 tokens), and FMR by length of the full match, with the corresponding propensity scores. A span is a repetition loop when, after collapsing whitespace, one unit of at most 200 characters repeats at least three times with at least 90\% of characters equal one period later. Loops are tested on spans of at least 20 tokens. When no generation has a full match of the required length, the bootstrap interval is $[0,0]$.

\begin{table*}[t]
\centering
\scriptsize
\setlength{\tabcolsep}{4pt}
\begin{tabular}{@{}lllcccc@{}}
\toprule
\textbf{Model} & \textbf{Data} & \textbf{Prompt} & \textbf{R20} & \textbf{R50} & \textbf{NVR50} & \textbf{Loop-only} \\
\midrule
Comma & CP & Unconditional & 67.2\% [66.3, 68.1] & 17.7\% [16.9, 18.4] & 0.0116 [0.0106, 0.0126] & 10 of 1777 \\
 &  & Minimal cue & 44.2\% [43.3, 45.2] & 10.9\% [10.3, 11.5] & 0.0110 [0.0100, 0.0121] & 16 of 1103 \\
 &  & Generic & 7.7\% [7.0, 8.5] & 0.9\% [0.7, 1.2] & 0.0006 [0.0004, 0.0009] & 36 of 130 \\
 &  & Specific & 15.5\% [14.4, 16.7] & 2.1\% [1.8, 2.5] & 0.0011 [0.0008, 0.0014] & 15 of 225 \\
 &  & Prefix & 53.5\% [51.1, 55.9] & 13.6\% [12.0, 15.2] & 0.0099 [0.0074, 0.0125] & 52 of 1410 \\
\midrule
DFM & CP & Unconditional & 6.7\% [6.2, 7.2] & 1.9\% [1.7, 2.2] & 0.0023 [0.0018, 0.0029] & 4 of 198 \\
 &  & Minimal cue & 19.4\% [18.6, 20.1] & 6.5\% [6.0, 7.0] & 0.0086 [0.0075, 0.0096] & 8 of 655 \\
 &  & Generic & 9.4\% [8.6, 10.2] & 0.6\% [0.4, 0.8] & 0.0005 [0.0003, 0.0007] & 23 of 82 \\
 &  & Specific & 12.6\% [11.6, 13.7] & 1.4\% [1.1, 1.7] & 0.0007 [0.0005, 0.0010] & 24 of 163 \\
 &  & Prefix & 51.0\% [48.6, 53.4] & 11.4\% [9.9, 13.0] & 0.0088 [0.0063, 0.0114] & 45 of 1184 \\
\midrule
DFM & DW & Unconditional & 49.4\% [48.4, 50.4] & 11.0\% [10.4, 11.6] & 0.0093 [0.0084, 0.0101] & 2 of 1102 \\
 &  & Minimal cue & 33.9\% [33.0, 34.9] & 3.3\% [2.9, 3.6] & 0.0039 [0.0033, 0.0047] & 3 of 328 \\
 &  & Generic & 15.1\% [14.1, 16.0] & 0.4\% [0.3, 0.5] & 0.0001 [0.0000, 0.0002] & 8 of 45 \\
 &  & Specific & 21.4\% [19.9, 23.0] & 2.8\% [2.3, 3.4] & 0.0012 [0.0008, 0.0017] & 19 of 301 \\
 &  & Prefix & 41.6\% [39.0, 44.1] & 10.1\% [8.6, 11.7] & 0.0108 [0.0082, 0.0135] & 2 of 1010 \\
\midrule
OLMo~3 & D3 & Unconditional & 84.3\% [83.6, 85.0] & 41.4\% [40.4, 42.4] & 0.0111 [0.0103, 0.0118] & 34 of 4170 \\
 &  & Minimal cue & 39.4\% [38.4, 40.3] & 8.6\% [8.1, 9.2] & 0.0048 [0.0042, 0.0053] & 5 of 868 \\
 &  & Generic & 8.9\% [8.2, 9.7] & 0.8\% [0.5, 1.1] & 0.0005 [0.0002, 0.0010] & 17 of 94 \\
 &  & Specific & 20.4\% [19.2, 21.7] & 1.7\% [1.3, 2.1] & 0.0004 [0.0002, 0.0007] & 2 of 172 \\
 &  & Prefix & 46.0\% [43.8, 48.3] & 9.0\% [7.6, 10.5] & 0.0036 [0.0024, 0.0050] & 8 of 909 \\
\bottomrule
\end{tabular}
\caption{Long verbatim matches. R20/R50: share of generations whose longest training span has $\geq$20/$\geq$50 Llama-2 tokens. NVR50: NVR counting only documents retrieved through spans of $\geq$50 tokens. Spans that are repetition loops are not counted; the last column gives the generations whose $\geq$50-token spans were all loops. 95\% bootstrap intervals.}
\label{tab:app-long}
\end{table*}

\begin{table*}[t]
\centering
\scriptsize
\setlength{\tabcolsep}{4pt}
\begin{tabular}{@{}lllccc@{}}
\toprule
\textbf{Model} & \textbf{Data} & \textbf{Prompt} & \textbf{FMR} & \textbf{FMR20} & \textbf{FMR50} \\
\midrule
Comma & CP & Unconditional & 0.0018 [0.0010, 0.0027] & 0.0014 [0.0007, 0.0022] & 0.0012 [0.0006, 0.0019] \\
 &  & Minimal cue & 0.0035 [0.0024, 0.0047] & 0.0032 [0.0021, 0.0044] & 0.0030 [0.0020, 0.0041] \\
 &  & Generic & 0.0014 [0.0007, 0.0022] & 0.0001 [0.0000, 0.0003] & 0.0001 [0.0000, 0.0003] \\
 &  & Specific & 0.0017 [0.0009, 0.0026] & 0.0001 [0.0000, 0.0003] & 0.0001 [0.0000, 0.0003] \\
 &  & Prefix & 0.0056 [0.0029, 0.0090] & 0.0033 [0.0008, 0.0064] & 0.0032 [0.0008, 0.0063] \\
\midrule
DFM & CP & Unconditional & 0.0015 [0.0008, 0.0023] & 0.0002 [0.0000, 0.0005] & 0.0002 [0.0000, 0.0005] \\
 &  & Minimal cue & 0.0071 [0.0055, 0.0088] & 0.0015 [0.0008, 0.0023] & 0.0015 [0.0008, 0.0023] \\
 &  & Generic & 0.0057 [0.0042, 0.0073] & 0.0002 [0.0000, 0.0005] & 0.0000 [0.0000, 0.0000] \\
 &  & Specific & 0.0029 [0.0018, 0.0041] & 0.0002 [0.0000, 0.0005] & 0.0001 [0.0000, 0.0003] \\
 &  & Prefix & 0.0037 [0.0020, 0.0056] & 0.0014 [0.0003, 0.0029] & 0.0014 [0.0003, 0.0029] \\
\midrule
DFM & DW & Unconditional & 0.0287 [0.0254, 0.0321] & 0.0070 [0.0054, 0.0087] & 0.0018 [0.0010, 0.0027] \\
 &  & Minimal cue & 0.0208 [0.0181, 0.0236] & 0.0026 [0.0017, 0.0037] & 0.0009 [0.0004, 0.0016] \\
 &  & Generic & 0.0011 [0.0004, 0.0020] & 0.0000 [0.0000, 0.0000] & 0.0000 [0.0000, 0.0000] \\
 &  & Specific & 0.0050 [0.0032, 0.0072] & 0.0001 [0.0000, 0.0003] & 0.0000 [0.0000, 0.0000] \\
 &  & Prefix & 0.0367 [0.0290, 0.0451] & 0.0085 [0.0047, 0.0130] & 0.0052 [0.0021, 0.0090] \\
\midrule
OLMo~3 & D3 & Unconditional & 0.0007 [0.0002, 0.0013] & 0.0007 [0.0002, 0.0013] & 0.0007 [0.0002, 0.0013] \\
 &  & Minimal cue & 0.0040 [0.0028, 0.0052] & 0.0037 [0.0025, 0.0049] & 0.0036 [0.0025, 0.0048] \\
 &  & Generic & 0.0002 [0.0000, 0.0005] & 0.0001 [0.0000, 0.0003] & 0.0001 [0.0000, 0.0003] \\
 &  & Specific & 0.0006 [0.0001, 0.0014] & 0.0003 [0.0000, 0.0009] & 0.0003 [0.0000, 0.0009] \\
 &  & Prefix & 0.0029 [0.0009, 0.0057] & 0.0019 [0.0001, 0.0045] & 0.0019 [0.0001, 0.0045] \\
\bottomrule
\end{tabular}
\caption{Full match ratio by length of the matched generation: all full matches (FMR, as in Table~\ref{tab:main-metrics}) and only those of $\geq$20 and $\geq$50 Llama-2 tokens. 95\% bootstrap intervals.}
\label{tab:app-fmr-length}
\end{table*}

\begin{table*}[t]
\centering
\scriptsize
\setlength{\tabcolsep}{4pt}
\begin{tabular}{@{}lllcccc@{}}
\toprule
\textbf{Model} & \textbf{Data} & \textbf{Metric} & \textbf{Uncond.} & \textbf{Min.\ cue} & \textbf{Generic} & \textbf{Specific} \\
\midrule
Comma & CP & $\PM_{\mathrm{R50}}$ & 0.565 [0.536, 0.598] & 0.444 [0.413, 0.479] & 0.065 [0.048, 0.085] & 0.134 [0.112, 0.159] \\
 &  & $\PM_{\mathrm{NVR50}}$ & 0.540 [0.476, 0.612] & 0.528 [0.463, 0.601] & 0.057 [0.035, 0.088] & 0.101 [0.071, 0.140] \\
 &  & $\PM_{\mathrm{FMR20}}$ & 0.298 [0.140, 0.636] & 0.492 [0.309, 0.792] & 0.029 [0.000, 0.150] & 0.029 [0.000, 0.154] \\
 &  & $\PM_{\mathrm{FMR50}}$ & 0.273 [0.120, 0.625] & 0.484 [0.297, 0.800] & 0.030 [0.000, 0.166] & 0.030 [0.000, 0.167] \\
\midrule
DFM & CP & $\PM_{\mathrm{R50}}$ & 0.145 [0.123, 0.170] & 0.362 [0.329, 0.399] & 0.049 [0.036, 0.064] & 0.109 [0.087, 0.135] \\
 &  & $\PM_{\mathrm{NVR50}}$ & 0.208 [0.155, 0.278] & 0.494 [0.421, 0.579] & 0.054 [0.032, 0.085] & 0.077 [0.052, 0.113] \\
 &  & $\PM_{\mathrm{FMR20}}$ & 0.125 [0.000, 0.499] & 0.517 [0.289, 0.852] & 0.125 [0.000, 0.500] & 0.126 [0.000, 0.501] \\
 &  & $\PM_{\mathrm{FMR50}}$ & 0.125 [0.000, 0.499] & 0.517 [0.289, 0.852] & 0.000 [0.000, 0.000] & 0.067 [0.000, 0.335] \\
\midrule
DFM & DW & $\PM_{\mathrm{R50}}$ & 0.522 [0.482, 0.564] & 0.244 [0.211, 0.280] & 0.035 [0.023, 0.050] & 0.220 [0.180, 0.264] \\
 &  & $\PM_{\mathrm{NVR50}}$ & 0.462 [0.400, 0.532] & 0.268 [0.214, 0.334] & 0.011 [0.002, 0.024] & 0.103 [0.068, 0.146] \\
 &  & $\PM_{\mathrm{FMR20}}$ & 0.452 [0.333, 0.607] & 0.234 [0.142, 0.380] & 0.000 [0.000, 0.000] & 0.012 [0.000, 0.045] \\
 &  & $\PM_{\mathrm{FMR50}}$ & 0.257 [0.136, 0.476] & 0.148 [0.057, 0.333] & 0.000 [0.000, 0.000] & 0.000 [0.000, 0.000] \\
\midrule
OLMo~3 & D3 & $\PM_{\mathrm{R50}}$ & 0.821 [0.798, 0.844] & 0.489 [0.448, 0.532] & 0.079 [0.054, 0.108] & 0.159 [0.122, 0.199] \\
 &  & $\PM_{\mathrm{NVR50}}$ & 0.756 [0.686, 0.824] & 0.571 [0.481, 0.670] & 0.119 [0.045, 0.239] & 0.107 [0.053, 0.183] \\
 &  & $\PM_{\mathrm{FMR20}}$ & 0.269 [0.083, 0.889] & 0.660 [0.428, 0.976] & 0.050 [0.000, 0.500] & 0.136 [0.000, 0.750] \\
 &  & $\PM_{\mathrm{FMR50}}$ & 0.269 [0.083, 0.889] & 0.654 [0.422, 0.975] & 0.050 [0.000, 0.500] & 0.136 [0.000, 0.750] \\
\bottomrule
\end{tabular}
\caption{Propensity scores on long matches and length-filtered full matches, against the prefix setting. 95\% bootstrap intervals.}
\label{tab:app-pm-long}
\end{table*}

\begin{figure*}[t]
  \centering
  \includegraphics[width=\textwidth]{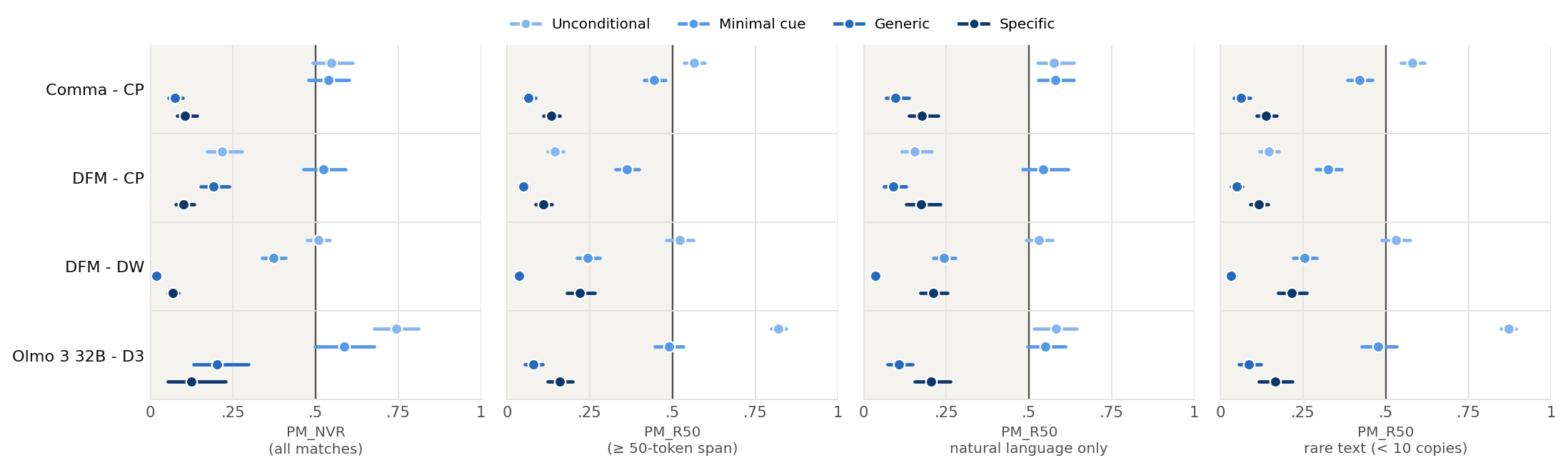}
  \caption{Propensity scores under stricter definitions, from left to right: $\PM_{\mathrm{NVR}}$ on all matches; $\PM_{\mathrm{R50}}$, counting only generations with a span of at least 50 tokens; $\PM_{\mathrm{R50}}$ on natural-language generations only; and $\PM_{\mathrm{R50}}$ on spans whose first 50 tokens occur fewer than 10 times in the training data. Each step removes one explanation other than memorization: short chance matches, code and boilerplate, and duplicated text. 95\% bootstrap intervals.}
  \label{fig:app-pm-robust}
\end{figure*}

\paragraph{Natural language only.} 
Figure~\ref{fig:app-pm-robust} shows how the propensity scores change as short matches, code, and duplicated text are removed in turn.
A generation is code-like if at least three lines, and at least 30\% of its lines, look like code (imports, declarations, braces, semicolons, comments), or if fewer than half of its characters are letters. Tables~\ref{tab:app-natural} and~\ref{tab:app-pm-natural} recompute the metrics without code-like generations.

\begin{table*}[t]
\centering
\scriptsize
\setlength{\tabcolsep}{4pt}
\begin{tabular}{@{}lllcccc@{}}
\toprule
\textbf{Model} & \textbf{Data} & \textbf{Prompt} & \textbf{Code-like} & \textbf{NVR} & \textbf{ALS} & \textbf{R50} \\
\midrule
Comma & CP & Unconditional & 29.2\% & 0.0290 [0.0262, 0.0319] & 29.82 [29.20, 30.47] & 10.7\% [10.0, 11.5] \\
 &  & Minimal cue & 8.9\% & 0.0366 [0.0338, 0.0395] & 28.87 [28.20, 29.57] & 10.9\% [10.2, 11.5] \\
 &  & Generic & 4.9\% & 0.0023 [0.0018, 0.0028] & 13.54 [13.21, 13.93] & 0.8\% [0.6, 1.1] \\
 &  & Specific & 5.1\% & 0.0033 [0.0025, 0.0041] & 15.29 [14.93, 15.68] & 1.7\% [1.4, 2.0] \\
 &  & Prefix & 47.1\% & 0.0203 [0.0140, 0.0276] & 24.70 [22.76, 26.85] & 7.9\% [6.2, 9.7] \\
\midrule
DFM & CP & Unconditional & 4.0\% & 0.0053 [0.0040, 0.0067] & 8.97 [8.77, 9.18] & 1.0\% [0.8, 1.2] \\
 &  & Minimal cue & 8.2\% & 0.0316 [0.0287, 0.0345] & 18.06 [17.47, 18.65] & 6.5\% [6.0, 7.0] \\
 &  & Generic & 5.6\% & 0.0063 [0.0054, 0.0074] & 13.06 [12.79, 13.35] & 0.5\% [0.4, 0.7] \\
 &  & Specific & 5.4\% & 0.0027 [0.0021, 0.0034] & 14.06 [13.71, 14.44] & 1.1\% [0.9, 1.4] \\
 &  & Prefix & 47.0\% & 0.0169 [0.0109, 0.0238] & 21.80 [20.26, 23.51] & 5.5\% [4.1, 7.0] \\
\midrule
DFM & DW & Unconditional & 4.0\% & 0.0512 [0.0485, 0.0539] & 27.00 [26.55, 27.44] & 11.5\% [10.8, 12.1] \\
 &  & Minimal cue & 0.7\% & 0.0276 [0.0255, 0.0297] & 20.25 [19.98, 20.53] & 3.3\% [2.9, 3.6] \\
 &  & Generic & 0.5\% & 0.0008 [0.0005, 0.0012] & 15.51 [15.32, 15.70] & 0.4\% [0.2, 0.5] \\
 &  & Specific & 1.4\% & 0.0033 [0.0025, 0.0041] & 16.67 [16.08, 17.27] & 2.7\% [2.2, 3.2] \\
 &  & Prefix & 2.0\% & 0.0464 [0.0406, 0.0527] & 25.56 [24.17, 27.04] & 10.2\% [8.7, 11.8] \\
\midrule
OLMo~3 & D3 & Unconditional & 82.3\% & 0.0207 [0.0163, 0.0255] & 26.34 [24.51, 28.29] & 8.7\% [7.4, 10.1] \\
 &  & Minimal cue & 10.9\% & 0.0116 [0.0103, 0.0129] & 25.44 [24.78, 26.14] & 7.7\% [7.1, 8.2] \\
 &  & Generic & 0.6\% & 0.0021 [0.0014, 0.0031] & 15.02 [14.77, 15.31] & 0.7\% [0.5, 1.0] \\
 &  & Specific & 0.8\% & 0.0011 [0.0005, 0.0022] & 17.38 [16.96, 17.84] & 1.6\% [1.2, 2.0] \\
 &  & Prefix & 16.0\% & 0.0075 [0.0044, 0.0114] & 24.16 [22.87, 25.58] & 6.3\% [5.0, 7.7] \\
\bottomrule
\end{tabular}
\caption{Metrics recomputed on natural-language generations only. Code-like: share of generations removed by the code heuristic (Appendix~\ref{sec:app-long-matches}). 95\% bootstrap intervals.}
\label{tab:app-natural}
\end{table*}

\begin{table*}[t]
\centering
\scriptsize
\setlength{\tabcolsep}{4pt}
\begin{tabular}{@{}lllcccc@{}}
\toprule
\textbf{Model} & \textbf{Data} & \textbf{Metric (natural language)} & \textbf{Uncond.} & \textbf{Min.\ cue} & \textbf{Generic} & \textbf{Specific} \\
\midrule
Comma & CP & $\PM_{\mathrm{NVR}}$ & 0.588 [0.508, 0.679] & 0.643 [0.568, 0.725] & 0.100 [0.070, 0.146] & 0.139 [0.098, 0.199] \\
 &  & $\PM_{\mathrm{R50}}$ & 0.576 [0.523, 0.634] & 0.579 [0.526, 0.636] & 0.095 [0.066, 0.133] & 0.175 [0.138, 0.222] \\
\midrule
DFM & CP & $\PM_{\mathrm{NVR}}$ & 0.239 [0.168, 0.340] & 0.652 [0.566, 0.744] & 0.273 [0.203, 0.372] & 0.140 [0.096, 0.206] \\
 &  & $\PM_{\mathrm{R50}}$ & 0.155 [0.117, 0.205] & 0.543 [0.479, 0.618] & 0.090 [0.062, 0.128] & 0.172 [0.128, 0.231] \\
\midrule
DFM & DW & $\PM_{\mathrm{NVR}}$ & 0.524 [0.490, 0.560] & 0.373 [0.339, 0.409] & 0.017 [0.010, 0.026] & 0.066 [0.050, 0.084] \\
 &  & $\PM_{\mathrm{R50}}$ & 0.530 [0.490, 0.571] & 0.242 [0.210, 0.278] & 0.034 [0.022, 0.048] & 0.210 [0.170, 0.253] \\
\midrule
OLMo~3 & D3 & $\PM_{\mathrm{NVR}}$ & 0.733 [0.628, 0.830] & 0.606 [0.499, 0.727] & 0.220 [0.135, 0.350] & 0.130 [0.051, 0.258] \\
 &  & $\PM_{\mathrm{R50}}$ & 0.581 [0.518, 0.645] & 0.550 [0.497, 0.608] & 0.106 [0.072, 0.149] & 0.203 [0.155, 0.260] \\
\bottomrule
\end{tabular}
\caption{Propensity scores on natural-language generations only. 95\% bootstrap intervals.}
\label{tab:app-pm-natural}
\end{table*}

\begin{figure*}[t]
  \centering
  \includegraphics[width=0.75\textwidth]{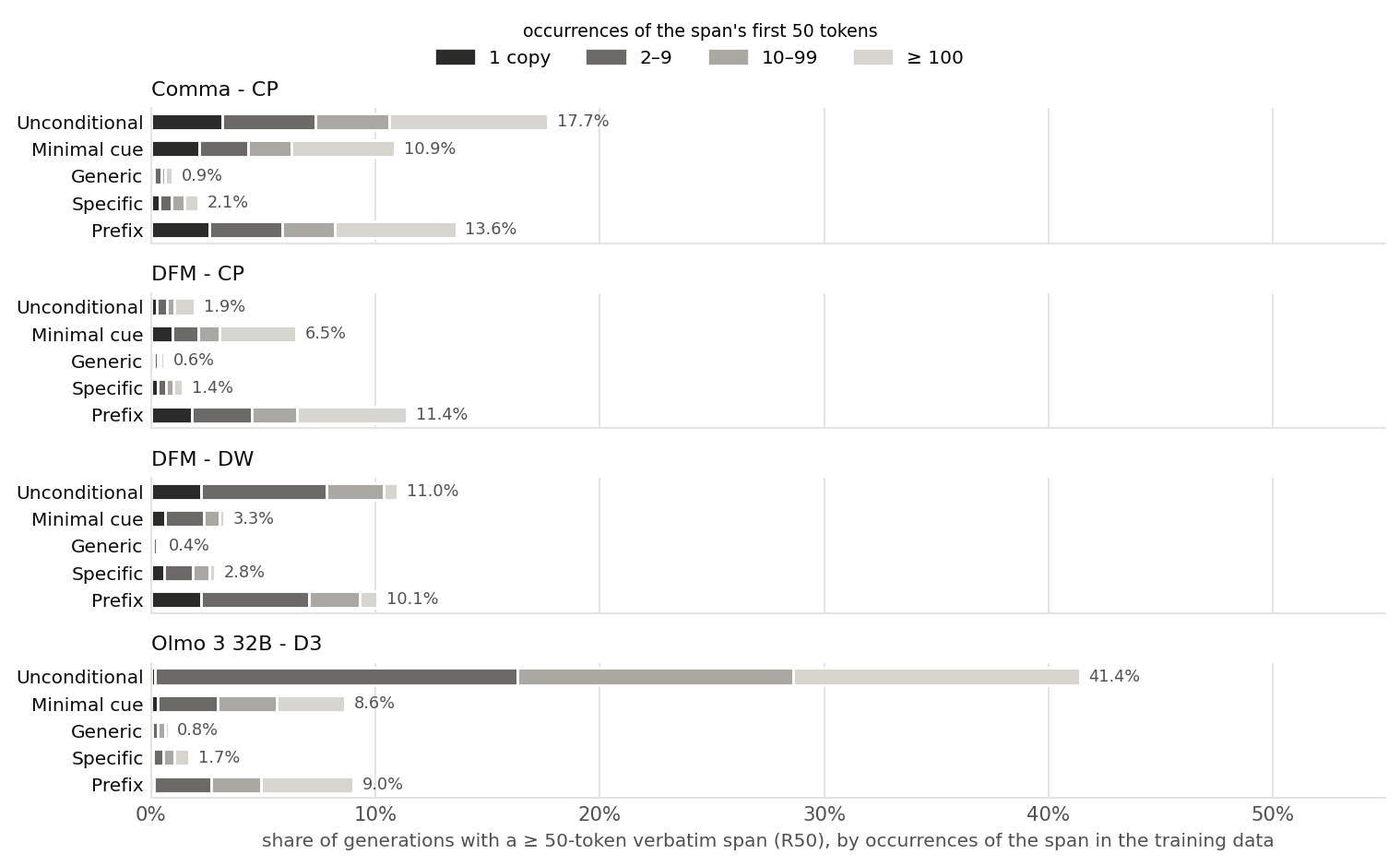}
  \caption{R50 per setting, split by how often the first 50 tokens of each generation's longest span occur in the training data (darker: rarer). In Dolma~3, copies are often adjacent near-duplicate documents, so single copies are rare in every setting.}
  \label{fig:app-duplication}
\end{figure*}

\paragraph{Duplication.} 
Figure~\ref{fig:app-duplication} shows R50 split by the number of occurrences.
For each generation with a span of at least 50 tokens, we count the occurrences of the span's first 50 tokens in the training index with infini-gram. Spans that start or end mid-word are trimmed by up to five tokens at each end to match the index tokenization. Counts are occurrences, not documents. In Dolma~3, the copies of a span are often adjacent near-duplicate documents, so a single copy is rare in every setting (Tables~\ref{tab:app-dup}, \ref{tab:app-rare}, and~\ref{tab:app-pm-rare}). With fewer than 10 copies, OLMo~3's unconditional R50 is 30.0\%, against 4.4\% for prefix attacks. On natural-language generations only, prompt-free $\PM_{\mathrm{R50}}$ is 0.576 and 0.579 for Comma, 0.530 for DFM Decoder on Dynaword (unconditional), and 0.581 and 0.550 for OLMo~3, although 82\% of OLMo~3's unconditional generations are code-like.

\begin{table*}[t]
\centering
\scriptsize
\setlength{\tabcolsep}{4pt}
\begin{tabular}{@{}lllccccccc@{}}
\toprule
\textbf{Model} & \textbf{Data} & \textbf{Prompt} & \textbf{$n$} & \textbf{Median} & \textbf{1} & \textbf{2--9} & \textbf{10--99} & \textbf{$\geq$100} & \textbf{Once (whole)} \\
\midrule
Comma & CP & Unconditional & 1767 & 25 & 18\% & 24\% & 18\% & 40\% & 50\% \\
 &  & Minimal cue & 1087 & 28 & 20\% & 20\% & 18\% & 42\% & 42\% \\
 &  & Generic & 94 & 10 & 14\% & 35\% & 17\% & 34\% & 31\% \\
 &  & Specific & 210 & 14 & 18\% & 25\% & 28\% & 29\% & 39\% \\
 &  & Prefix & 1358 & 25 & 19\% & 24\% & 17\% & 40\% & 44\% \\
\midrule
DFM & CP & Unconditional & 194 & 66 & 14\% & 23\% & 16\% & 47\% & 54\% \\
 &  & Minimal cue & 647 & 149 & 15\% & 18\% & 14\% & 53\% & 45\% \\
 &  & Generic & 59 & 11 & 19\% & 29\% & 24\% & 29\% & 41\% \\
 &  & Specific & 139 & 12 & 20\% & 26\% & 25\% & 29\% & 37\% \\
 &  & Prefix & 1139 & 36 & 16\% & 24\% & 18\% & 43\% & 42\% \\
\midrule
DFM & DW & Unconditional & 1100 & 4 & 20\% & 51\% & 23\% & 6\% & 52\% \\
 &  & Minimal cue & 325 & 4 & 19\% & 53\% & 21\% & 7\% & 44\% \\
 &  & Generic & 37 & 5 & 24\% & 41\% & 19\% & 16\% & 57\% \\
 &  & Specific & 282 & 5 & 21\% & 44\% & 26\% & 9\% & 53\% \\
 &  & Prefix & 1008 & 5 & 22\% & 48\% & 22\% & 8\% & 50\% \\
\midrule
OLMo~3 & D3 & Unconditional & 4136 & 18 & 0\% & 39\% & 30\% & 31\% & 1\% \\
 &  & Minimal cue & 862 & 26 & 3\% & 31\% & 31\% & 35\% & 7\% \\
 &  & Generic & 77 & 21 & 6\% & 34\% & 39\% & 21\% & 10\% \\
 &  & Specific & 170 & 36 & 6\% & 25\% & 29\% & 40\% & 17\% \\
 &  & Prefix & 901 & 71 & 1\% & 28\% & 24\% & 46\% & 5\% \\
\bottomrule
\end{tabular}
\caption{Duplication of the longest $\geq$50-token span of each generation: number of generations with such a span, median number of occurrences of its first 50 tokens in the training index, distribution of these occurrences, and share whose whole span occurs once.}
\label{tab:app-dup}
\end{table*}

\begin{table}[t]
\centering
\scriptsize
\setlength{\tabcolsep}{4pt}
\begin{tabular}{@{}lllcc@{}}
\toprule
\textbf{Model} & \textbf{Data} & \textbf{Prompt} & \textbf{$<$10 copies} & \textbf{1 copy} \\
\midrule
Comma & CP & Unconditional & 12.0\% [11.4, 12.6] & 6.6\% [6.1, 7.1] \\
 &  & Minimal cue & 6.3\% [5.8, 6.8] & 3.6\% [3.2, 4.0] \\
 &  & Generic & 0.6\% [0.4, 0.8] & 0.3\% [0.1, 0.5] \\
 &  & Specific & 1.4\% [1.1, 1.7] & 0.8\% [0.6, 1.0] \\
 &  & Prefix & 8.7\% [7.6, 9.9] & 5.1\% [4.3, 5.9] \\
\midrule
DFM & CP & Unconditional & 1.2\% [1.0, 1.4] & 0.7\% [0.5, 0.8] \\
 &  & Minimal cue & 3.4\% [3.1, 3.8] & 2.1\% [1.8, 2.3] \\
 &  & Generic & 0.4\% [0.2, 0.5] & 0.2\% [0.1, 0.3] \\
 &  & Specific & 0.9\% [0.7, 1.1] & 0.5\% [0.3, 0.6] \\
 &  & Prefix & 7.1\% [6.1, 8.1] & 3.9\% [3.2, 4.6] \\
\midrule
DFM & DW & Unconditional & 9.7\% [9.1, 10.3] & 4.6\% [4.2, 5.0] \\
 &  & Minimal cue & 2.9\% [2.6, 3.3] & 1.2\% [1.0, 1.4] \\
 &  & Generic & 0.3\% [0.2, 0.4] & 0.1\% [0.1, 0.2] \\
 &  & Specific & 2.4\% [1.9, 2.8] & 1.1\% [0.8, 1.4] \\
 &  & Prefix & 8.6\% [7.3, 10.0] & 3.8\% [3.1, 4.5] \\
\midrule
OLMo~3 & D3 & Unconditional & 30.0\% [29.1, 30.9] & 0.4\% [0.3, 0.5] \\
 &  & Minimal cue & 4.0\% [3.7, 4.4] & 0.5\% [0.4, 0.6] \\
 &  & Generic & 0.4\% [0.3, 0.6] & 0.1\% [0.0, 0.1] \\
 &  & Specific & 0.9\% [0.6, 1.2] & 0.3\% [0.2, 0.4] \\
 &  & Prefix & 4.4\% [3.6, 5.3] & 0.4\% [0.2, 0.7] \\
\bottomrule
\end{tabular}
\caption{R50 restricted to spans whose first 50 tokens occur fewer than 10 times or exactly once in the training data. 95\% bootstrap intervals.}
\label{tab:app-rare}
\end{table}

\begin{table*}[t]
\centering
\scriptsize
\setlength{\tabcolsep}{4pt}
\begin{tabular}{@{}lllcccc@{}}
\toprule
\textbf{Model} & \textbf{Data} & \textbf{Metric} & \textbf{Uncond.} & \textbf{Min.\ cue} & \textbf{Generic} & \textbf{Specific} \\
\midrule
Comma & CP & $\PM_{\mathrm{R50}}$ ($<$ 10 copies) & 0.580 [0.546, 0.615] & 0.420 [0.384, 0.459] & 0.062 [0.041, 0.087] & 0.138 [0.111, 0.167] \\
 &  & $\PM_{\mathrm{R50}}$ (1 copy) & 0.567 [0.525, 0.610] & 0.417 [0.373, 0.463] & 0.049 [0.024, 0.083] & 0.132 [0.102, 0.166] \\
\midrule
DFM & CP & $\PM_{\mathrm{R50}}$ ($<$ 10 copies) & 0.146 [0.119, 0.176] & 0.326 [0.288, 0.367] & 0.048 [0.032, 0.067] & 0.116 [0.089, 0.146] \\
 &  & $\PM_{\mathrm{R50}}$ (1 copy) & 0.145 [0.110, 0.184] & 0.345 [0.297, 0.397] & 0.042 [0.022, 0.065] & 0.106 [0.074, 0.144] \\
\midrule
DFM & DW & $\PM_{\mathrm{R50}}$ ($<$ 10 copies) & 0.530 [0.489, 0.573] & 0.253 [0.219, 0.292] & 0.033 [0.021, 0.047] & 0.215 [0.174, 0.260] \\
 &  & $\PM_{\mathrm{R50}}$ (1 copy) & 0.547 [0.496, 0.601] & 0.237 [0.193, 0.287] & 0.033 [0.015, 0.055] & 0.224 [0.169, 0.285] \\
\midrule
OLMo~3 & D3 & $\PM_{\mathrm{R50}}$ ($<$ 10 copies) & 0.871 [0.849, 0.893] & 0.477 [0.425, 0.533] & 0.087 [0.057, 0.123] & 0.166 [0.117, 0.222] \\
 &  & $\PM_{\mathrm{R50}}$ (1 copy) & 0.482 [0.353, 0.641] & 0.532 [0.404, 0.684] & 0.154 [0.061, 0.294] & 0.389 [0.244, 0.560] \\
\bottomrule
\end{tabular}
\caption{Propensity scores of R50 restricted to rare spans. 95\% bootstrap intervals.}
\label{tab:app-pm-rare}
\end{table*}

\paragraph{Repeated sampling and targeted extraction.} Table~\ref{tab:app-repeated} gives, for the prompted settings, the share of prompts with a long match in at least one of their 10 generations. Tables~\ref{tab:app-discoverable} and~\ref{tab:app-discoverable-dup} give the rate of targeted extraction in the prefix setting: a generation counts when its next 50 tokens equal the next 50 tokens of its source document, after whitespace normalization.

\begin{table*}[t]
\centering
\scriptsize
\setlength{\tabcolsep}{4pt}
\begin{tabular}{@{}lllccc@{}}
\toprule
\textbf{Model} & \textbf{Data} & \textbf{Prompt} & \textbf{Full match $\geq$20 in $\geq$1/10} & \textbf{Mean} & \textbf{$\geq$50-token span in $\geq$1/10} \\
\midrule
Comma & CP & Generic & 0.1\% [0.0, 0.3] & 1.0 & 7.2\% [5.7, 8.8] \\
 &  & Specific & 0.1\% [0.0, 0.3] & 1.0 & 16.2\% [13.9, 18.5] \\
 &  & Prefix & 0.9\% [0.4, 1.5] & 3.7 & 35.9\% [32.9, 38.9] \\
\midrule
DFM & CP & Generic & 0.2\% [0.0, 0.5] & 1.0 & 5.5\% [4.1, 7.0] \\
 &  & Specific & 0.2\% [0.0, 0.5] & 1.0 & 11.1\% [9.2, 13.1] \\
 &  & Prefix & 0.5\% [0.1, 1.0] & 2.8 & 31.4\% [28.5, 34.3] \\
\midrule
DFM & DW & Generic & 0.0\% [0.0, 0.0] & 0.0 & 3.3\% [2.2, 4.5] \\
 &  & Specific & 0.1\% [0.0, 0.3] & 1.0 & 15.5\% [13.3, 17.8] \\
 &  & Prefix & 2.7\% [1.7, 3.8] & 3.1 & 21.1\% [18.6, 23.7] \\
\midrule
OLMo~3 & D3 & Generic & 0.1\% [0.0, 0.3] & 1.0 & 5.9\% [4.5, 7.4] \\
 &  & Specific & 0.1\% [0.0, 0.3] & 3.0 & 10.7\% [8.8, 12.7] \\
 &  & Prefix & 0.4\% [0.1, 0.8] & 4.8 & 22.2\% [19.6, 24.8] \\
\bottomrule
\end{tabular}
\caption{Share of prompts with a full match of $\geq$20 tokens, or a $\geq$50-token span, in at least one of their 10 generations; ``Mean of 10'': in how many of the 10 generations it occurs, when it occurs. 95\% bootstrap intervals.}
\label{tab:app-repeated}
\end{table*}

\begin{table}[t]
\centering
\scriptsize
\setlength{\tabcolsep}{4pt}
\begin{tabular}{@{}llcccc@{}}
\toprule
\textbf{Model} & \textbf{Data} & \textbf{Gens.} & \textbf{Per gen.} & \textbf{$\geq$1 of 10} & \textbf{Mean} \\
\midrule
Comma & CP & 9988 & 2.15\% [1.35, 3.01] & 3.1\% [2.0, 4.2] & 6.9 \\
\midrule
DFM & CP & 9982 & 1.93\% [1.17, 2.76] & 2.9\% [1.9, 4.0] & 6.7 \\
\midrule
DFM & DW & 9996 & 1.04\% [0.55, 1.62] & 2.5\% [1.6, 3.5] & 4.2 \\
\midrule
OLMo~3 & D3 & 9988 & 1.28\% [0.67, 1.98] & 1.8\% [1.0, 2.7] & 7.1 \\
\bottomrule
\end{tabular}
\caption{Targeted extraction in the prefix setting: share of generations that reproduce the true next 50 Llama-2 tokens of their source document (after whitespace normalization), per generation and in at least one of 10. 95\% bootstrap intervals.}
\label{tab:app-discoverable}
\end{table}

\begin{table}[t]
\centering
\scriptsize
\setlength{\tabcolsep}{4pt}
\begin{tabular}{@{}llccc@{}}
\toprule
\textbf{Model} & \textbf{Data} & \textbf{1} & \textbf{2--9} & \textbf{$\geq$10} \\
\midrule
Comma & CP & 0.9\% (938) & 1.1\% (36) & 48.5\% (26) \\
\midrule
DFM & CP & 0.8\% (938) & 0.6\% (36) & 45.0\% (26) \\
\midrule
DFM & DW & 0.6\% (941) & 4.4\% (57) & 100.0\% (2) \\
\midrule
OLMo~3 & D3 & 0.2\% (86) & 0.9\% (703) & 2.8\% (211) \\
\bottomrule
\end{tabular}
\caption{Per-generation targeted extraction rate, grouped by how often the 100-token source sequence (prefix and true continuation) occurs in the training data (number of prompts in parentheses).}
\label{tab:app-discoverable-dup}
\end{table}

\section{Detailed Results of Main Experiments}
\label{sec:app-main-tables}
Tables~\ref{tab:main-metrics} and~\ref{tab:main-propensity} give the values plotted in Figures~\ref{fig:main-metrics} and~\ref{fig:propensity-main-metrics}, with 95\% bootstrap intervals.

\begin{table*}[t]
\centering
\scriptsize
\setlength{\tabcolsep}{4pt}
\renewcommand{\arraystretch}{1.15}
\begin{tabular}{@{}llccccc@{}}
\toprule
\textbf{Model / Data} & \textbf{Metric} & \textbf{Uncond.} & \textbf{Min.\ cue} & \textbf{Generic} & \textbf{Specific} & \textbf{Prefix} \\
\midrule
\multirow{3}{*}{Comma / CP} & NVR & \shortstack{0.0353\\{\scriptsize[0.0327, 0.0379]}} & \shortstack{0.0338\\{\scriptsize[0.0313, 0.0365]}} & \shortstack{0.0023$^{\downarrow}$\\{\scriptsize[0.0018, 0.0028]}} & \shortstack{0.0034$^{\downarrow}$\\{\scriptsize[0.0027, 0.0042]}} & \shortstack{0.0291\\{\scriptsize[0.0226, 0.0363]}} \\
 & FMR & \shortstack{0.0018$^{\downarrow}$\\{\scriptsize[0.0010, 0.0027]}} & \shortstack{0.0035\\{\scriptsize[0.0024, 0.0047]}} & \shortstack{0.0014$^{\downarrow}$\\{\scriptsize[0.0007, 0.0022]}} & \shortstack{0.0017$^{\downarrow}$\\{\scriptsize[0.0009, 0.0027]}} & \shortstack{0.0056\\{\scriptsize[0.0028, 0.0089]}} \\
 & ALS & \shortstack{35.76$^{\uparrow}$\\{\scriptsize[35.08, 36.42]}} & \shortstack{29.10$^{\downarrow}$\\{\scriptsize[28.47, 29.76]}} & \shortstack{13.88$^{\downarrow}$\\{\scriptsize[13.53, 14.27]}} & \shortstack{16.23$^{\downarrow}$\\{\scriptsize[15.78, 16.71]}} & \shortstack{32.46\\{\scriptsize[30.60, 34.42]}} \\
\midrule
\multirow{3}{*}{DFM / CP} & NVR & \shortstack{0.0072$^{\downarrow}$\\{\scriptsize[0.0058, 0.0087]}} & \shortstack{0.0286\\{\scriptsize[0.0260, 0.0313]}} & \shortstack{0.0061$^{\downarrow}$\\{\scriptsize[0.0052, 0.0071]}} & \shortstack{0.0029$^{\downarrow}$\\{\scriptsize[0.0023, 0.0035]}} & \shortstack{0.0260\\{\scriptsize[0.0202, 0.0325]}} \\
 & FMR & \shortstack{0.0015$^{\downarrow}$\\{\scriptsize[0.0008, 0.0023]}} & \shortstack{0.0071$^{\uparrow}$\\{\scriptsize[0.0055, 0.0087]}} & \shortstack{0.0057\\{\scriptsize[0.0042, 0.0073]}} & \shortstack{0.0029\\{\scriptsize[0.0018, 0.0041]}} & \shortstack{0.0037\\{\scriptsize[0.0020, 0.0056]}} \\
 & ALS & \shortstack{10.31$^{\downarrow}$\\{\scriptsize[10.03, 10.60]}} & \shortstack{18.58$^{\downarrow}$\\{\scriptsize[18.04, 19.14]}} & \shortstack{13.38$^{\downarrow}$\\{\scriptsize[13.10, 13.67]}} & \shortstack{15.19$^{\downarrow}$\\{\scriptsize[14.70, 15.70]}} & \shortstack{30.24\\{\scriptsize[28.55, 32.01]}} \\
\midrule
\multirow{3}{*}{DFM / DW} & NVR & \shortstack{0.0475\\{\scriptsize[0.0450, 0.0501]}} & \shortstack{0.0273$^{\downarrow}$\\{\scriptsize[0.0253, 0.0294]}} & \shortstack{0.0008$^{\downarrow}$\\{\scriptsize[0.0005, 0.0012]}} & \shortstack{0.0034$^{\downarrow}$\\{\scriptsize[0.0026, 0.0043]}} & \shortstack{0.0460\\{\scriptsize[0.0402, 0.0520]}} \\
 & FMR & \shortstack{0.0287\\{\scriptsize[0.0255, 0.0320]}} & \shortstack{0.0208$^{\downarrow}$\\{\scriptsize[0.0180, 0.0237]}} & \shortstack{0.0011$^{\downarrow}$\\{\scriptsize[0.0004, 0.0020]}} & \shortstack{0.0050$^{\downarrow}$\\{\scriptsize[0.0032, 0.0072]}} & \shortstack{0.0367\\{\scriptsize[0.0287, 0.0451]}} \\
 & ALS & \shortstack{26.25\\{\scriptsize[25.82, 26.69]}} & \shortstack{20.25$^{\downarrow}$\\{\scriptsize[19.98, 20.53]}} & \shortstack{15.53$^{\downarrow}$\\{\scriptsize[15.34, 15.72]}} & \shortstack{17.29$^{\downarrow}$\\{\scriptsize[16.62, 18.00]}} & \shortstack{25.44\\{\scriptsize[24.05, 26.89]}} \\
\midrule
\multirow{3}{*}{OLMo~3 / D3} & NVR & \shortstack{0.0249$^{\uparrow}$\\{\scriptsize[0.0236, 0.0263]}} & \shortstack{0.0121\\{\scriptsize[0.0109, 0.0133]}} & \shortstack{0.0021$^{\downarrow}$\\{\scriptsize[0.0015, 0.0031]}} & \shortstack{0.0012$^{\downarrow}$\\{\scriptsize[0.0005, 0.0023]}} & \shortstack{0.0085\\{\scriptsize[0.0058, 0.0120]}} \\
 & FMR & \shortstack{0.0007$^{\downarrow}$\\{\scriptsize[0.0002, 0.0013]}} & \shortstack{0.0040\\{\scriptsize[0.0028, 0.0053]}} & \shortstack{0.0002$^{\downarrow}$\\{\scriptsize[0.0000, 0.0005]}} & \shortstack{0.0006$^{\downarrow}$\\{\scriptsize[0.0001, 0.0014]}} & \shortstack{0.0029\\{\scriptsize[0.0009, 0.0058]}} \\
 & ALS & \shortstack{53.42$^{\uparrow}$\\{\scriptsize[52.62, 54.24]}} & \shortstack{26.54\\{\scriptsize[25.90, 27.19]}} & \shortstack{15.10$^{\downarrow}$\\{\scriptsize[14.83, 15.41]}} & \shortstack{17.68$^{\downarrow}$\\{\scriptsize[17.19, 18.23]}} & \shortstack{26.90\\{\scriptsize[25.48, 28.39]}} \\
\bottomrule
\end{tabular}
\caption{Memorization metrics per model, training corpus, and setting (the values of Figure~\ref{fig:main-metrics}), with 95\% bootstrap confidence intervals. NVR: near-verbatim recall; FMR: full match ratio; ALS: average longest span (Llama-2 tokens). $^{\uparrow}$/$^{\downarrow}$: the difference from the prefix setting of the same model and corpus is significantly above/below 0 (95\% bootstrap interval). Degenerate outputs are excluded. CP: Common Pile, DW: Dynaword, D3: Dolma~3.}
\label{tab:main-metrics}
\end{table*}

\begin{table*}[t]
\centering
\scriptsize
\setlength{\tabcolsep}{6pt}
\renewcommand{\arraystretch}{1.15}
\begin{tabular}{@{}llcccc@{}}
\toprule
& & \textbf{Uncond.} & \textbf{Min.\ cue} & \textbf{Generic} & \textbf{Specific} \\
\midrule
\multirow{2}{*}{Comma / CP} & $\PM_{\mathrm{NVR}}$ & \shortstack{\textbf{0.548}\\{\scriptsize[0.490, 0.614]}} & \shortstack{\textbf{0.538}\\{\scriptsize[0.480, 0.602]}} & \shortstack{0.073\\{\scriptsize[0.054, 0.100]}} & \shortstack{0.105\\{\scriptsize[0.078, 0.141]}} \\
 & $\PM_{\mathrm{FMR}}$ & \shortstack{0.243\\{\scriptsize[0.134, 0.413]}} & \shortstack{0.384\\{\scriptsize[0.255, 0.560]}} & \shortstack{0.200\\{\scriptsize[0.101, 0.353]}} & \shortstack{0.233\\{\scriptsize[0.118, 0.400]}} \\
\midrule
\multirow{2}{*}{DFM / CP} & $\PM_{\mathrm{NVR}}$ & \shortstack{0.217\\{\scriptsize[0.169, 0.277]}} & \shortstack{\textbf{0.523}\\{\scriptsize[0.464, 0.591]}} & \shortstack{0.190\\{\scriptsize[0.151, 0.240]}} & \shortstack{0.100\\{\scriptsize[0.075, 0.134]}} \\
 & $\PM_{\mathrm{FMR}}$ & \shortstack{0.288\\{\scriptsize[0.158, 0.459]}} & \shortstack{\textbf{0.657}\\{\scriptsize[0.538, 0.787]}} & \shortstack{\textbf{0.606}\\{\scriptsize[0.481, 0.747]}} & \shortstack{0.441\\{\scriptsize[0.294, 0.615]}} \\
\midrule
\multirow{2}{*}{DFM / DW} & $\PM_{\mathrm{NVR}}$ & \shortstack{\textbf{0.508}\\{\scriptsize[0.473, 0.543]}} & \shortstack{0.372\\{\scriptsize[0.338, 0.410]}} & \shortstack{0.017\\{\scriptsize[0.010, 0.026]}} & \shortstack{0.068\\{\scriptsize[0.052, 0.087]}} \\
 & $\PM_{\mathrm{FMR}}$ & \shortstack{0.439\\{\scriptsize[0.382, 0.505]}} & \shortstack{0.362\\{\scriptsize[0.306, 0.428]}} & \shortstack{0.029\\{\scriptsize[0.011, 0.055]}} & \shortstack{0.121\\{\scriptsize[0.078, 0.175]}} \\
\midrule
\multirow{2}{*}{OLMo~3 / D3} & $\PM_{\mathrm{NVR}}$ & \shortstack{\textbf{0.744}\\{\scriptsize[0.674, 0.811]}} & \shortstack{\textbf{0.586}\\{\scriptsize[0.497, 0.679]}} & \shortstack{0.201\\{\scriptsize[0.130, 0.298]}} & \shortstack{0.124\\{\scriptsize[0.054, 0.229]}} \\
 & $\PM_{\mathrm{FMR}}$ & \shortstack{0.194\\{\scriptsize[0.064, 0.467]}} & \shortstack{\textbf{0.579}\\{\scriptsize[0.396, 0.820]}} & \shortstack{0.064\\{\scriptsize[0.000, 0.235]}} & \shortstack{0.171\\{\scriptsize[0.023, 0.454]}} \\
\bottomrule
\end{tabular}
\caption{Propensity scores $\PM=f^p/(f^p+f^c)$ of each non-adversarial setting against the prefix setting (the values of Figure~\ref{fig:propensity-main-metrics}), with 95\% bootstrap confidence intervals. 0.5 is neutral; values $\geq 0.5$ are in bold. Degenerate outputs are excluded.}
\label{tab:main-propensity}
\end{table*}

\section{Continual Pre-Training and Training Stages}
\label{sec:app-stages}
\begin{figure*}[t]
  \centering
  \includegraphics[width=0.85\textwidth]{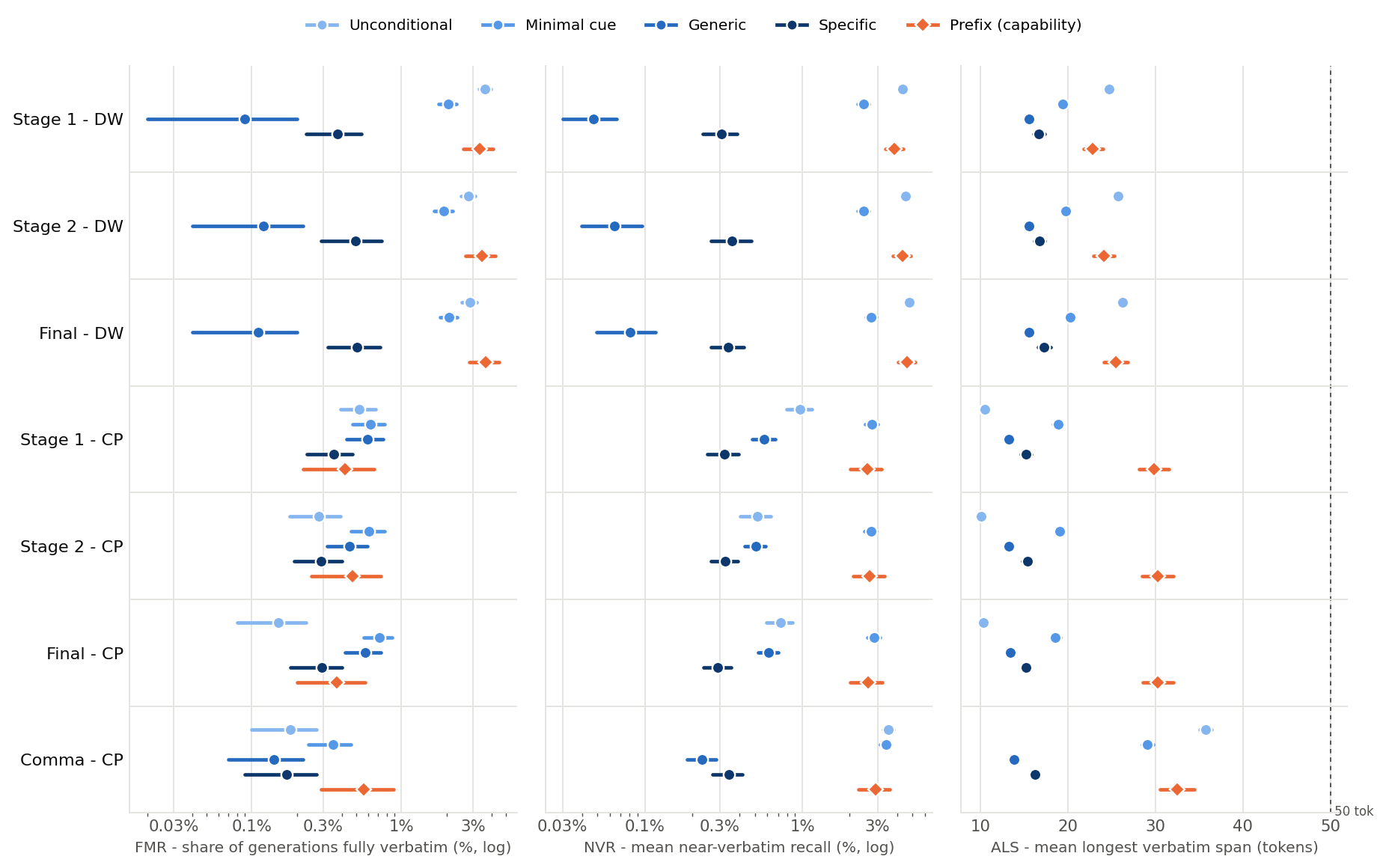}
  \caption{Memorization of DFM Decoder's three training stages (Stage~1, Stage~2, and the final checkpoint) on Dynaword (DW) and Common Pile (CP), and of Comma, its base model, on Common Pile, with 95\% bootstrap intervals. Axes as in Figure~\ref{fig:main-metrics}.}
  \label{fig:app-stages-metrics}
\end{figure*}

\begin{figure*}[t]
  \centering
  \includegraphics[width=\textwidth]{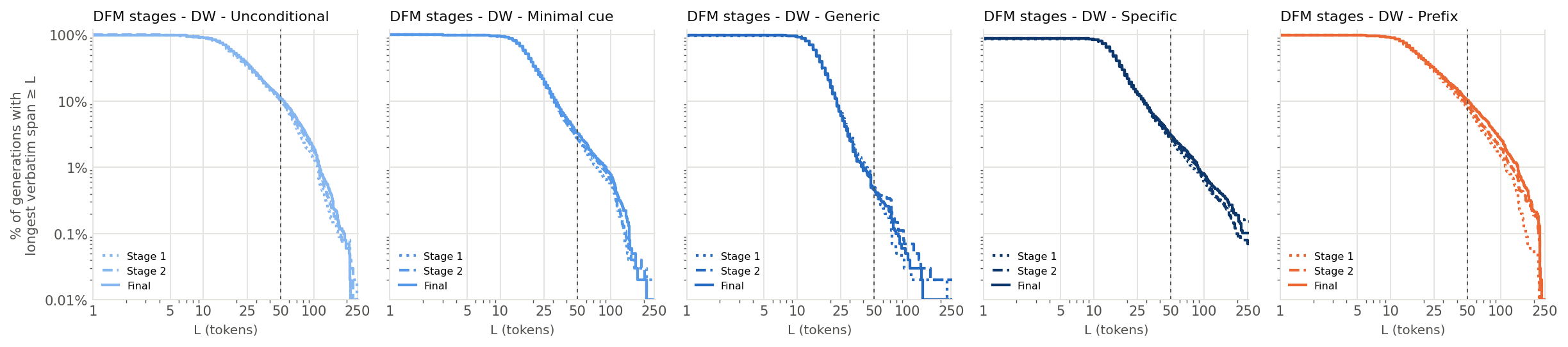}\\[4pt]
  \includegraphics[width=\textwidth]{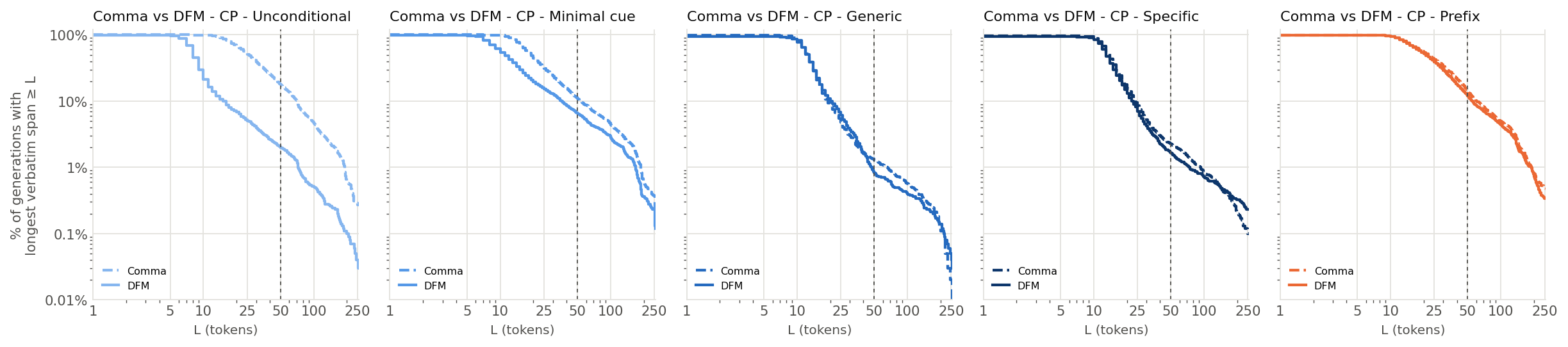}
  \caption{Longest-span survival curves per setting (log--log, as in Figure~\ref{fig:app-survival}). \textbf{Top:} DFM Decoder on Dynaword at Stage~1 (dotted), Stage~2 (dashed), and the final checkpoint (solid). \textbf{Bottom:} Comma (dashed) and DFM Decoder (solid) on Common Pile, on the same prompts.}
  \label{fig:app-survival-continual}
\end{figure*}

Figures~\ref{fig:app-stages-metrics} and~\ref{fig:app-survival-continual} show the same comparisons as plots.
Table~\ref{tab:app-continual} gives the differences between DFM Decoder and Comma on Common Pile. Tables~\ref{tab:app-stages-dw} and~\ref{tab:app-stages-cp} give the metrics of DFM Decoder's three training stages. On Common Pile, 6 of the 30 stage--final differences are significant and show no consistent direction. On Dynaword, prefix NVR and ALS grow with each stage (NVR 0.038, 0.043, 0.046; ALS 22.8, 24.1, 25.4 tokens), consistent with continued training on Dynaword.

\begin{table*}[t]
\centering
\scriptsize
\setlength{\tabcolsep}{4pt}
\begin{tabular}{@{}lcccc@{}}
\toprule
\textbf{Prompt} & \textbf{$\Delta$NVR} & \textbf{$\Delta$FMR} & \textbf{$\Delta$ALS} & \textbf{Resampling} \\
\midrule
Unconditional & $-$0.0280 [$-$0.0311, $-$0.0251] $\downarrow$ & $-$0.0003 [$-$0.0014, +0.0009] $\approx$ & $-$25.45 [$-$26.15, $-$24.74] $\downarrow$ & independent \\
Minimal cue & $-$0.0053 [$-$0.0080, $-$0.0026] $\downarrow$ & +0.0036 [+0.0017, +0.0055] $\uparrow$ & $-$10.52 [$-$11.12, $-$9.92] $\downarrow$ & paired \\
Generic & +0.0038 [+0.0029, +0.0048] $\uparrow$ & +0.0043 [+0.0028, +0.0058] $\uparrow$ & $-$0.50 [$-$0.94, $-$0.08] $\downarrow$ & paired \\
Specific & $-$0.0005 [$-$0.0013, +0.0003] $\approx$ & +0.0012 [+0.0000, +0.0024] $\uparrow$ & $-$1.04 [$-$1.58, $-$0.50] $\downarrow$ & paired \\
Prefix & $-$0.0030 [$-$0.0056, $-$0.0005] $\downarrow$ & $-$0.0019 [$-$0.0044, +0.0001] $\approx$ & $-$2.22 [$-$3.06, $-$1.42] $\downarrow$ & paired \\
\bottomrule
\end{tabular}
\caption{Differences DFM Decoder $-$ Comma on Common Pile, per setting, with 95\% bootstrap intervals (paired on shared prompts; unconditional generations resampled independently). $\uparrow$/$\downarrow$/$\approx$: interval above/below/including 0.}
\label{tab:app-continual}
\end{table*}

\begin{table*}[t]
\centering
\scriptsize
\setlength{\tabcolsep}{4pt}
\begin{tabular}{@{}llccccc@{}}
\toprule
\textbf{Prompt} & \textbf{Stage} & \textbf{NVR} & \textbf{FMR} & \textbf{ALS} & \textbf{$\PM_{\mathrm{NVR}}$} & \textbf{$\PM_{\mathrm{FMR}}$} \\
\midrule
Unconditional & Stage 1 & 0.0430 [0.0406, 0.0455] & 0.0363 [0.0327, 0.0399] & 24.73 [24.34, 25.13] & 0.529 [0.493, 0.566] & 0.520 [0.462, 0.585] \\
\midrule
Unconditional & Stage 2 & 0.0451 [0.0426, 0.0476] & 0.0280 [0.0248, 0.0313] & 25.72 [25.31, 26.14] & 0.511 [0.476, 0.548] & 0.447 [0.388, 0.513] \\
\midrule
Unconditional & Final & 0.0475 [0.0450, 0.0501] & 0.0287 [0.0255, 0.0320] & 26.25 [25.82, 26.69] & 0.508 [0.473, 0.543] & 0.439 [0.382, 0.505] \\
\midrule
Minimal cue & Stage 1 & 0.0245 [0.0225, 0.0265] & 0.0206 [0.0178, 0.0234] & 19.43 [19.17, 19.71] & 0.390 [0.353, 0.429] & 0.381 [0.322, 0.448] \\
\midrule
Minimal cue & Stage 2 & 0.0244 [0.0225, 0.0265] & 0.0192 [0.0166, 0.0220] & 19.73 [19.46, 20.01] & 0.362 [0.326, 0.400] & 0.356 [0.301, 0.423] \\
\midrule
Minimal cue & Final & 0.0273 [0.0253, 0.0294] & 0.0208 [0.0180, 0.0237] & 20.25 [19.98, 20.53] & 0.372 [0.338, 0.410] & 0.362 [0.306, 0.428] \\
\midrule
Generic & Stage 1 & 0.0005 [0.0003, 0.0007] & 0.0009 [0.0002, 0.0019] & 15.52 [15.29, 15.76] & 0.012 [0.008, 0.018] & 0.026 [0.006, 0.057] \\
\midrule
Generic & Stage 2 & 0.0006 [0.0004, 0.0010] & 0.0012 [0.0005, 0.0022] & 15.55 [15.33, 15.78] & 0.015 [0.009, 0.022] & 0.033 [0.012, 0.062] \\
\midrule
Generic & Final & 0.0008 [0.0005, 0.0012] & 0.0011 [0.0004, 0.0020] & 15.53 [15.34, 15.72] & 0.017 [0.010, 0.026] & 0.029 [0.011, 0.055] \\
\midrule
Specific & Stage 1 & 0.0030 [0.0023, 0.0039] & 0.0037 [0.0022, 0.0054] & 16.68 [16.07, 17.30] & 0.074 [0.056, 0.095] & 0.100 [0.061, 0.149] \\
\midrule
Specific & Stage 2 & 0.0036 [0.0026, 0.0047] & 0.0049 [0.0029, 0.0073] & 16.75 [16.15, 17.35] & 0.077 [0.056, 0.102] & 0.124 [0.074, 0.186] \\
\midrule
Specific & Final & 0.0034 [0.0026, 0.0043] & 0.0050 [0.0032, 0.0072] & 17.29 [16.62, 18.00] & 0.068 [0.052, 0.087] & 0.121 [0.078, 0.175] \\
\midrule
Prefix & Stage 1 & 0.0383 [0.0333, 0.0436] & 0.0335 [0.0260, 0.0412] & 22.81 [21.73, 23.97] & — & — \\
\midrule
Prefix & Stage 2 & 0.0431 [0.0375, 0.0489] & 0.0347 [0.0269, 0.0427] & 24.10 [22.87, 25.37] & — & — \\
\midrule
Prefix & Final & 0.0460 [0.0402, 0.0520] & 0.0367 [0.0287, 0.0451] & 25.44 [24.05, 26.89] & — & — \\
\bottomrule
\end{tabular}
\caption{Memorization of Dynaword across the three training stages of DFM Decoder (Final is the released checkpoint), with 95\% bootstrap intervals.}
\label{tab:app-stages-dw}
\end{table*}

\begin{table*}[t]
\centering
\scriptsize
\setlength{\tabcolsep}{4pt}
\begin{tabular}{@{}llccccc@{}}
\toprule
\textbf{Prompt} & \textbf{Stage} & \textbf{NVR} & \textbf{FMR} & \textbf{ALS} & \textbf{$\PM_{\mathrm{NVR}}$} & \textbf{$\PM_{\mathrm{FMR}}$} \\
\midrule
Unconditional & Stage 1 & 0.0096 [0.0079, 0.0115] & 0.0052 [0.0038, 0.0067] & 10.48 [10.21, 10.77] & 0.271 [0.219, 0.336] & 0.553 [0.420, 0.709] \\
\midrule
Unconditional & Stage 2 & 0.0052 [0.0040, 0.0063] & 0.0028 [0.0018, 0.0039] & 10.06 [9.79, 10.34] & 0.161 [0.122, 0.211] & 0.373 [0.247, 0.542] \\
\midrule
Unconditional & Final & 0.0072 [0.0058, 0.0087] & 0.0015 [0.0008, 0.0023] & 10.31 [10.03, 10.60] & 0.217 [0.169, 0.277] & 0.288 [0.158, 0.459] \\
\midrule
Minimal cue & Stage 1 & 0.0275 [0.0250, 0.0301] & 0.0062 [0.0047, 0.0078] & 18.85 [18.32, 19.40] & 0.515 [0.456, 0.581] & 0.596 [0.464, 0.742] \\
\midrule
Minimal cue & Stage 2 & 0.0272 [0.0247, 0.0298] & 0.0061 [0.0046, 0.0077] & 19.05 [18.52, 19.61] & 0.503 [0.444, 0.571] & 0.564 [0.438, 0.713] \\
\midrule
Minimal cue & Final & 0.0286 [0.0260, 0.0313] & 0.0071 [0.0055, 0.0087] & 18.58 [18.04, 19.14] & 0.523 [0.464, 0.591] & 0.657 [0.538, 0.787] \\
\midrule
Generic & Stage 1 & 0.0057 [0.0048, 0.0068] & 0.0059 [0.0043, 0.0075] & 13.28 [12.99, 13.59] & 0.181 [0.143, 0.229] & 0.584 [0.452, 0.737] \\
\midrule
Generic & Stage 2 & 0.0051 [0.0043, 0.0059] & 0.0045 [0.0032, 0.0059] & 13.27 [12.98, 13.57] & 0.159 [0.125, 0.202] & 0.489 [0.356, 0.652] \\
\midrule
Generic & Final & 0.0061 [0.0052, 0.0071] & 0.0057 [0.0042, 0.0073] & 13.38 [13.10, 13.67] & 0.190 [0.151, 0.240] & 0.606 [0.481, 0.747] \\
\midrule
Specific & Stage 1 & 0.0032 [0.0025, 0.0040] & 0.0035 [0.0024, 0.0047] & 15.22 [14.65, 15.83] & 0.110 [0.081, 0.147] & 0.456 [0.316, 0.630] \\
\midrule
Specific & Stage 2 & 0.0032 [0.0026, 0.0039] & 0.0029 [0.0019, 0.0040] & 15.36 [14.82, 15.93] & 0.108 [0.081, 0.144] & 0.383 [0.254, 0.552] \\
\midrule
Specific & Final & 0.0029 [0.0023, 0.0035] & 0.0029 [0.0018, 0.0041] & 15.19 [14.70, 15.70] & 0.100 [0.075, 0.134] & 0.441 [0.294, 0.615] \\
\midrule
Prefix & Stage 1 & 0.0259 [0.0201, 0.0320] & 0.0042 [0.0022, 0.0066] & 29.79 [28.12, 31.55] & — & — \\
\midrule
Prefix & Stage 2 & 0.0268 [0.0208, 0.0333] & 0.0047 [0.0026, 0.0072] & 30.23 [28.50, 32.06] & — & — \\
\midrule
Prefix & Final & 0.0260 [0.0202, 0.0325] & 0.0037 [0.0020, 0.0056] & 30.24 [28.55, 32.01] & — & — \\
\bottomrule
\end{tabular}
\caption{Memorization of Common Pile across the three training stages of DFM Decoder (Final is the released checkpoint), with 95\% bootstrap intervals.}
\label{tab:app-stages-cp}
\end{table*}

\section{Inspected Generations}
\label{sec:app-inspection}
For each model, corpus, and setting, we inspected 100 matched generations: all full matches, or 100 drawn at random when there are more, followed by the generations with the longest partial matches. Degenerate outputs were excluded. Claude Opus 5.5 categorized the matched text of each generation automatically, using a code heuristic, regular expressions for license and copyright notices, a compression-based loop test, and reading of the generations, and the authors verified the categories. Because the partial matches are the longest ones, the counts in Table~\ref{tab:app-inspection} describe the longest matches of each setting, not all generations. Typical content per model:
(i) Comma: software licenses, code scaffolds, publisher front matter, dataset citation pages, wiki maintenance pages, and task instructions without a prompt; open-access license notices after specific prompts; Bible passages after generic prompts.
(ii) DFM Decoder on Dynaword: formulaic notices from 18th- and 19th-century newspapers (sales, rentals, probate summons, auctions).
(iii) OLMo~3: coding exercises and code without a prompt; problem statements, wiki pages, and reading-comprehension passages after a cue word; thesis-repository cover pages after specific prompts.

\begin{table}[t]
\centering
\scriptsize
\setlength{\tabcolsep}{3pt}
\begin{tabular}{@{}lllrrrrr@{}}
\toprule
\textbf{Model} & \textbf{Data} & \textbf{Setting} & \textbf{$n$} & \textbf{Loop} & \textbf{Code} & \textbf{License} & \textbf{Other} \\
\midrule
Comma & CP & Uncond. & 94 & 3 & 47 & 50 & 30 \\
 &  & Min.\ cue & 95 & 2 & 5 & 13 & 77 \\
 &  & Generic & 87 & 26 & 11 & 17 & 37 \\
 &  & Specific & 84 & 11 & 28 & 33 & 21 \\
 &  & Prefix & 76 & 15 & 51 & 28 & 9 \\
\midrule
DFM & CP & Uncond. & 87 & 3 & 42 & 16 & 40 \\
 &  & Min.\ cue & 44 & 1 & 2 & 1 & 40 \\
 &  & Generic & 43 & 17 & 6 & 4 & 20 \\
 &  & Specific & 72 & 20 & 29 & 13 & 22 \\
 &  & Prefix & 77 & 20 & 53 & 33 & 6 \\
\midrule
DFM & DW & Uncond. & 6 & 0 & 0 & 0 & 6 \\
 &  & Min.\ cue & 2 & 0 & 0 & 0 & 2 \\
 &  & Generic & 45 & 2 & 2 & 0 & 42 \\
 &  & Specific & 50 & 18 & 21 & 0 & 25 \\
 &  & Prefix & 16 & 0 & 0 & 0 & 16 \\
\midrule
OLMo~3 & D3 & Uncond. & 100 & 2 & 72 & 8 & 25 \\
 &  & Min.\ cue & 97 & 1 & 12 & 3 & 81 \\
 &  & Generic & 94 & 14 & 4 & 0 & 77 \\
 &  & Specific & 97 & 2 & 10 & 26 & 59 \\
 &  & Prefix & 90 & 0 & 21 & 12 & 69 \\
\bottomrule
\end{tabular}
\caption{Content of the inspected generations whose longest training span has at least 50 tokens ($n$, out of 100 inspected per setting). Loop: the matched text is a repetition loop; Code: code-like generation; License: license or copyright notice; Other: none of these, which still includes templated web text. Code and License can overlap. On Dynaword most inspected generations are short full matches, so few reach 50 tokens.}
\label{tab:app-inspection}
\end{table}

\section{Use of AI Assistants}
AI assistants were used to support coding, grammar and style revisions, and literature search. Claude Opus 5.5 was also used to categorize the inspected generations (Appendix~\ref{sec:app-inspection}); the authors verified them.

\end{document}